\documentclass[journal]{IEEEtran}
\ifCLASSINFOpdf
\else
\fi
\ifCLASSOPTIONcompsoc
\usepackage[caption=false,font=normalsize,labelfont=sf,textfont=sf]{subfig}
\else
\usepackage[caption=false,font=footnotesize]{subfig}
\fi
\usepackage{amssymb}%para las letras dobles
\usepackage{amsmath}
\usepackage{color}
\usepackage{graphicx}

\usepackage{epstopdf}

\usepackage{float}
\restylefloat{table}

\def\R{{\mathbb {R}}}
\def\sumi{\sum_{i=1}^n}
\def\sump{\sum_{p=1}^m}

\def\sumk{\sum_{k=1}^N}

\newtheorem{teo}{Theorem}[section]

\newtheorem{remark}[teo]{Remark}

\newtheorem{exmp}{Example}[section]

\begin{document}
%
% paper title
% Titles are generally capitalized except for words such as a, an, and, as,
% at, but, by, for, in, nor, of, on, or, the, to and up, which are usually
% not capitalized unless they are the first or last word of the title.
% Linebreaks \\ can be used within to get better formatting as desired.
% Do not put math or special symbols in the title.
\title{Super-resolution method for data deconvolution from a single acquisition by superposition of virtual point sources}
%
%
% author names and IEEE memberships
% note positions of commas and nonbreaking spaces ( ~ ) LaTeX will not break
% a structure at a ~ so this keeps an author's name from being broken across
% two lines.
% use \thanks{} to gain access to the first footnote area
% a separate \thanks must be used for each paragraph as LaTeX2e's \thanks
% was not built to handle multiple paragraphs
%

\author{Sandra Mart\'inez
        %\IEEEmembership{Fellow,~OSA,}
        and Oscar E. Mart\'inez~%\IEEEmembership{Life~Fellow,~IEEE}% <-this % stops a space
        \thanks{Sandra Martinez is with Departamento de Matemática, FCEyN-UBA, IMAS, CONICET,
	Buenos Aires, Argentina. e-mail: smartin@dm.uba.ar.
	}
\thanks{Oscar Martinez is with Departamento de Física, FI-UBA, Photonics Lab,
	CONICET, Buenos Aires, Argentina.e-mail: omartinez@fi.uba.ar.}% <-this % stops a space
% <-this % stops a space

\thanks{The work was partially financed by grant PICT 2015-1523 from the Agencia Nacional de Promoción Científica y Tecnológica and grant  ubacyt2014 20920160100624BA  from the Universidad de Buenos Aires}

\thanks{Manuscript version March 2018.}}

\maketitle

% As a general rule, do not put math, special symbols or citations
% in the abstract or keywords.
\begin{abstract}
In this work we present a new method for data deconvolution from a single acquisition without a sparsity prior, that allows the retrieval of the target function with super-resolution. The measured data are fit by a superposition of virtual point sources (SUPPOSe) of equal intensity. The cloud of virtual point sources approximates the actual distribution of sources that can be discrete or continuous. In this manner only the positions of the sources need to be determined by an algorithm, that minimizes the norm of the difference between the measured data and the convolution of the superposed point sources with the Instrument Response Function. An upper bound for the uncertainty in the position of the sources was derived and two very different experimental situations were used for the test (an optical spectrum and fluorescent microscopy images) showing excellent reconstructions and agreement with the predicted uncertainties, achieving $\lambda/10$ resolution for the microscope and a fivefold improvement in the spectral resolution for the spectrometer. The method also provides a way to determine the optimum number of sources to be used for the fit. 
\end{abstract}

% Note that keywords are not normally used for peerreview papers.
\begin{IEEEkeywords}
super-resolution, data deconvolution, signal processing algorithms, signal resolution, image resolution.
\end{IEEEkeywords}

% For peer review papers, you can put extra information on the cover
% page as needed:
% \ifCLASSOPTIONpeerreview
% \begin{center} \bfseries EDICS Category: 3-BBND \end{center}
% \fi
%
% For peerreview papers, this IEEEtran command inserts a page break and
% creates the second title. It will be ignored for other modes.
\IEEEpeerreviewmaketitle

\section{Introduction}
% The very first letter is a 2 line initial drop letter followed
% by the rest of the first word in caps.
% 
% form to use if the first word consists of a single letter:
% \IEEEPARstart{A}{demo} file is ....
% 
% form to use if you need the single drop letter followed by
% normal text (unknown if ever used by the IEEE):
% \IEEEPARstart{A}{}demo file is ....
% 
% Some journals put the first two words in caps:
% \IEEEPARstart{T}{his demo} file is ....
% 
% Here we have the typical use of a "T" for an initial drop letter
% and "HIS" in caps to complete the first word.
\IEEEPARstart{A}{ll} measurements are blurred and distorted by the Instrument response function (IRF) also called Point Spread Function (PSF) in imaging and Impulse Response in the time domain. This distortion can arise from physical limitations such as limited bandwidth of the instrument (time response  or diffraction in the case of spatial measurements), from instrument aberrations, blurring from moving samples, aliasing from discrete sampling (pixel size, sampling interval) and noise. In many cases the relation between the target information $R(x)$ and the measured data  $S(x)$ are given by a convolution with the IRF ($I$) i.e:
\begin{equation}\label{partida}S(x)=R*I(x)+\eta(x)+B(x).\end{equation}

 Here $R(x)$ is assumed to be contained  in a ball in  $\R^D$ and the measurement  $S(x)$ samples a region contained in that ball. 
  The function $I$ is obtained after pixelation the original IRF $J$, that is;  $I(x)=J *K_p$ and  $K_p$ is the characteristic function of the cube $Q_p=[-\frac{d_p}{2}, \frac{d_p}{2}]^D$.
  The function $I(x)$ for any practical purpose can be assumed to have compact support and  the measurement $S(x)$ will sample a ball larger than the resulting support off $R*I$.
     
Also $S$ is only sampled for certain values  $\{x_i\}_{i=1}^n$, which 
 are the pixels,  being each $x_i$ a vector in $\R^D$, $n$  the number of pixels with $x_{i+1}-x_{i}\in Q_p$.  
Finally $\eta$ is a random variable that averages to zero representing the noise  and $B$ is the background (its noise is included in $\eta$).

Very efficient strategies have been developed to overcome the limitations from the undersampling (aliasing) such as resolution enhancement (also called super-resolution or high resolution image reconstruction, \cite{Park}). The technique relies in overcoming the aliasing arising from the undersampling by taking multiple displaced records (or images) of the data (assumed invariant). This scheme obviously requires multiple images for the reconstruction. 

Many deblurring algorithms have been developed to restore images distorted by moving targets or out of focus acquisitions \cite{Campisi} but the main point in this cases is that the camera is not loosing high spatial frequency components, the image is simply distorted (blurred) due to phase distortions. Hence this type of reconstructions do not correspond strictly to super-resolution techniques (recovery of attenuated high frequency components).

 A fantastic solution known as compressed sensing has been found for sparse data showing that for certain set of undersampled functions an exact recovery is possible (see  \cite{Donoho} and \cite{Candes}). This idea of compressed sensing was used recently also in \cite{Candes2} to recover $R$ from $S$, when $I$ is given by a theoretical and particular function and $B=0$. The sparcity prior is a very restrictive constraint as in \cite{Candes2} it was shown that the quality of the reconstruction is severely hurt when three or more sources are overlapping within the IRF. Recent works have extended the technique to sparse signal restoration on a continuous grid   (see \cite{Chouzenoux,Duval} \cite{Ekanadham} \cite{Zhu}). But the reconstruction of continuous distribution or high density of sources has not been possible.

 The conceptual limit to sparsity is having a single point source (a molecule for an image, a spectral line, an extremely short pulse, etc.) and finding its location by solving a least square problem from \eqref{partida}. This localization scheme has been successfully used for particle tracking in microscopy (see \cite{Thompson}) and more recently for super-resolution imaging by STORM or PALM (see \cite{Betzig, Hess,Rust}). The techniques rely in successively locate fluorescent molecules one at a time until the total image is reconstructed. As the localization method requires that two molecules are not simultaneously ignited with overlapping IRF, this requires the acquisition of thousands or even tens of thousands of images for a single reconstruction. Using compressed sensing schemes STORM (see (\cite{Min,Zhu2012}) or locating simultaneously several sources (see \cite{Huang}) image acquisition has been speeded by localizing simultaneously several molecules within the point spread function. Still we are dealing with extremely sparse individual images requiring hundreds or thousands of images to complete the restoration. The super-resolution recovery from inversion of the convolution equation presented in \eqref{partida} from a single frame still remains a challenge.

Several strategies have been followed with the simplification of assuming that the instrument response function has translational invariance. For this case the inversion of the problem given in \eqref{partida} can be done by trivial Fourier analysis in the absence of noise, but that requires special care because the noise is amplified in such simple minded processing. 

One example of the limitation given by \eqref{partida} is the deconvolution of microscopic images obtained from fluorescent samples (typical in biology). In this cases 3 dimensional image reconstruction from stacks of images at different planes or confocal scans where performed following different strategies. The simplest case is truncated inverse filtering, that is,  deconvolving in the Fourier transform space by dividing by the Fourier transformation of the PSF truncated to avoid the noise amplification at high frequencies (see  \cite{McNally})  or Wigner filtering (see \cite{Tommasi}  and \cite{Erhardt}). None of these techniques can recover the high frequency components of the image, and hence do not provide a super-resolution restoration. A way to deal with the noise has been to convert the deconvolution problem in a linear least square fit problem by searching for the target function $R$ that convolved with the instrument response function minimizes the distance to the measured data $S$ (see \cite{Tikhonov}) but these methods do not recover the high frequency components remaining limited by the instrument function response cut-off. In addition negative values for $R$ frequently are obtained and they are very sensitive to the quality of the measurement of the IRF.

The key to overcome the IRF limit is to incorporate additional information to the process. One first approach is to force $R\geq0$ and performing a nonlinear least square fit (see \cite{Carrington}), at the expense of a high computational cost. An alternative is to clip the negative values in an iterative algorithm as done in the Tikhonov-Miller algorithm (see  \cite{Kempen}, \cite{Verveer} and \cite{Voort}). More robust deconvolution schemes have been obtained adding wavelet denoising methods (see \cite{Monvel}) but only marginal increase in the resolution is obtained.

One key aspect of any deconvolution technique is the quality of the IRF used. Examples of the effort to measure the IRF are  \cite{Hearn} for spectrometers, for x-ray diffractometers (see \cite{Gozzo} \cite{Jacobs}), and \cite{Mboula} for image PSF for Astronomy where the compressed sensing schemes are incorporated. For fluorescence microscopy an alternative approach has been to compute theoretically the PSF for an objective based on the manufacturer objective parameters (see \cite{Gibson}). But the actual IRF will not be in general shift invariant, will have aberrations not accounted by theoretical predictions and must be accurately measured for high resolution deconvolution methods.

%In this work we present a new algorithm for data deconvolution that allows the retrieval of the target function $R$ with super-resolution with a simple approach that after a precise measurement of the IRF, the measured data are fit by a superposition of point sources (SUPPOSe) of equal intensity. In this manner only the positions of the sources need to be determined by an  algorithm that minimizes the norm of the difference between the measured data and the convolution of the superposed point sources with the IRF. 
%
%In the next section the fundamentals of the method are presented, followed by the estimation of the uncertainties of the reconstruction and finally two experimental examples are presented, optical spectra (one dimensional problem with subtracted background) and fluorescent microscope images (two dimensional example with unknown background).

 A common experimental scenario as encountered when the detectors measure intensity (light, X-ray, particles, etc.) is that $R$ is  positive. For such cases in this work we present a new algorithm for data deconvolution that allows the retrieval of the target function $R$ with super-resolution with a simple approach of assuming that the source distribution can be approximated by a superposition of virtual point sources of equal intensities. The function $R$ can be either a continuous distribution or a discrete superposition of sources of arbitrary intensities. These virtual sources reconstruct the actual distribution by locating them in such a manner that the cloud of sources reproduces with enough precision the actual distribution. In this manner only the positions of the sources need to be determined by an algorithm, that minimizes the norm of the difference between the measured data and the convolution of the superposed point sources with the IRF. The problem of finding the intensity of each position is converted to finding the position of the point sources. The intensity fit is achieved by accumulating many sources in close proximity. The positions have no constraint, and the reconstruction is made from a single acquisition or image. The sparsity prior is not required although it will be shown that the method has a better resolution for sparser sources.  
 
 In the next section the fundamentals of the method are presented, followed by the estimation of the uncertainties of the reconstruction and a determination of the number of sources needed for maximal resolution and finally an artificially synthesized fluorescent image and two experimental examples are presented, one is an optical spectra (one dimensional problem with subtracted background) and the other one is fluorescent microscope images (two dimensional example with unknown background). For the fluorescent images the simulated data allow the comparison of the retrieved solution with the ground truth, verifying the predicted resolution.

%SUPerposition of POint SourcEs= SUPPOSE

\section{Description of the method}\label{method}

The method we propose is to approximate the target function $R$, that we wish to measure with better resolution than that given by the instrument response function, by a superposition of virtual point sources of identical intensities $\alpha$ so that the only unknown are the positions of the sources. Hence the approximate solution $\tilde{R}(x)$ would result:
\begin{equation}\label{Ansatz}\tilde{R}(x)= \alpha \sum_{k=1}^{N}\delta(x-{\tilde{a}_k)},\end{equation}
here for each $k=1,...,N$,  $\tilde{a}_k\in \R^D$ and can be repeated. In this manner the intensity at a given point is adjusted by placing more particles at that location. It is important to notice that the sources do not pretend to locate the position of actual sources and reproduce their intensity. The target function $R$ is approximated by a cloud of identical virtual sources and the target function $R$ can be either discrete or a continuous distribution.

We will use the notation  with tilde to indicate an approximate solution of the same variable without tilde. $N$ is the number of point sources used for the fit and ${\tilde{a}_k}$ are the positions of the point sources. 
We will define for the presentation of the result a new pixel that we will call superpixel as it will express the measurement with super-resolution. As the acquisition of the data are oversampled (the pixel size is much smaller than the IRF width) the superpixel can be smaller or larger than the original pixel, and hence we will not call it subpixel as usually done for super-resolution.

 \subsection{Case with no background}\label{no background}
We will start the discussion for the case the background in \eqref{partida} does not exist or can be substracted (the noise arrising from the substraction is absorbed in $\eta$). In this case the recorded signal $S$ can be reconstructed approximately by \begin{equation}\label{Stildefondocero}\tilde{S}(x)=\tilde{R}*\tilde{I}(x)=\alpha \sum_{k=1}^N \tilde{I}(x-\tilde{a}_k).\end{equation}
 
Given  $N$ and $\alpha$, we search for the position of the point sources that yield a minimum of

\begin{equation}\label{chimin1}
\chi^2= \| S-\tilde{S}\|^2=\sum_{i=1}^n (S(x_i)-\tilde{S}(x_i))^2.
\end{equation} 

Here $\tilde{I}$ is some approximation of the IRF function $I$. This  $\tilde{I}$ is obtained, in practice  by fitting by an adequate function the results of several measurements of a calibration source that is assumed point-like (see Apendix \ref{fitting}).

Hence the goal is to find $\{\tilde{a_k}^j\}$, the minimum of \eqref{chimin1}  with  $j=1,...,D$ and $k=1,...,N$. Remember that $D$ is the dimension of the space and $N$ is the number of point sources used for the fit, chosen as described later.

 We want to mention here that we chose the $2-$ norm because this will allow us to estimate the  uncertainties in the positions used to determine the optimum value for $N$. Other norms can be used and might even yield better reconstructions for specific cases, but the prediction of the precision of the reconstruction would be difficult.

\begin{remark}
Observe that in this case by \eqref{Stildefondocero}, we have
$$\sum_{i=1}^n  \tilde{S}(x_i)=\alpha \sum_{k=1}^N  \sum_{i=1}^n \tilde{I}(x_i-a_k).$$
Since in the case where the function $\tilde{I}$ is invariant under translations and the pixel is small we have 
$$\sum_{i=1}^n \tilde{I}(x_i-a_k)\sim\sum_{i=1}^n \tilde{I}(x_i),$$
and since we expect $\tilde{S}\sim S$
we chose if $I$ is normalized, 
$$\alpha=\frac{\sum_{i=1}^n S(x_i)}{N }.$$
\end{remark} 	

 \subsection*{Algorithm}
\label{sec.algo}

In this subsection we describe briefly the steps of the algorithm.

In the next section we will define the  parameters $\sigma_{op}$ (optimal accuracy $\sigma$ in the positions of the sources) and $N_{op}$ (number of sources $N$ that optimizes the accuracy in the position ). These optimal parameters are bounds of the original ones. They cannot be calculated a priori (because they depend on $R$), so we developed an algorithm to approximate this parameters and subsequently arrive to the desired solution.

\begin{enumerate}
%\item Start with some arbitrary $N$ for example we can choose $\alpha_0=\frac{\max(S)}{\mbox{Coef} \tilde{I}(0)}$ . Then we use the relation $\sumi S(x_i)=\alpha N$ to determine $N$. Coef is arbitrarily chosen and it would be the number of point sources needed if that maximum arises from a single point source.
\item Start with some arbitrary $N$ for example we can choose an initial value for $\alpha_0$ and then we use the relation $\sumi S(x_i)=\alpha_0 N$ to determine $N$. 
\item Then we use a Genetic Algorithm to solve the Unconstrained Minimization problem. We found the genetic algorithm adequate for our examples as the large dimension of the problem hinders from using optimization methods that converge to local minima. Other global optimization methods can be used if found convenient.

\item Make an histogram of the solution vector $\{\tilde{a_k}\}$ for different bins $d_{bin}=d_p,\frac{d_p}  {2}\frac{d_p}{4},...$ where $d_p$ is the pixel size and define $m_{bin}$ the number of non-zero bins.  Now we have an approximation of $y_p$, $R_p$ and $m$ so we can compute all the terms involved in $N_{op}$ and $\alpha$ is scaled accordingly. Return to step (2).

\item We finally choose $d_s=d_{bin}=\sigma_{op}$. With this process we do not choose a priori which is the superpixel, this is part of the calculation. The bounds of $\sigma_{op}$ depends on the measurement of the function $I$, the noise, etc.

\item Convolve the obtained point sources with the known shape of the point source used for the determination of the IRF. This gives a continuous solution and there is no need to define a superpixel. This step is optional.
\end{enumerate}

\begin{remark}\label{ventajas}
Observe that one of the main advantages of this method is that  
  we  are dealing with a minimization problem in $\R^N\times \R^D$ without any  constraints  nor assumptions on the sparcity of the problem.
The size of the superpixel  is not fixed a priori, so the positions of the $\{\tilde{a}^j_k\}$ are all free. Finally, we can choose if plotting using a superpixel defined by the resoution of the method or create a signal (image, spectrum, etc.) convolving the point sources with a distributed source the same size as the source used to measure the IRF.
\end{remark}
\begin{remark}
Observe that the Genetic Algorithm cannot guaranty that the solution is a global minimum  $\{\tilde{a}^j_k\}$. The algorithm stops when $\chi^2$ is small enough compared to the bounds described in the next section. Simulations with synthesized data with similar structure to that of the problem to be solved are necessary to gain confidence in the solution obtained.  
\end{remark}
 \subsection{Case with  background}\label{background}

If  the background in \eqref{partida}  is constant and unknown 
 we call
 $S_{dev}=S-\frac{1}{n}\sum_{i=1}^n S(x_i) $,   $\tilde{I}_{dev}=\tilde{I}-\frac{1}{n}\sum_{i=1}^n \tilde{I}(x_i)$ and 
\begin{equation}\label{Stildefondo}\tilde{S}(x)=\tilde{R}*\tilde{I}_{dev}(x)=\alpha \sum_{k=1}^N \tilde{I}_{dev}(x-\tilde{a}_k)……….\end{equation}

Here, given  $N$ we find for  $k=1,...,N$  $\tilde{a}_k\in \R^D,$  and $\alpha$ such that minimizes:

\begin{equation}\label{chimin2}
\chi^2= \|S_{dev}-\tilde{S}\|=\sum_{i=1}^n (S_{dev}(x_i)-\tilde{S}(x_i))^2.
\end{equation}

 Here we are using that all the random variables $\eta(x_i)$ are independent and have the same  distribution. Then the mean over all the pixels  it is equal to $\frac{1}{n}\sum_{i=1}^n \eta(x_i)=0$. Also we are using that the background is constant.

In this case the algorithm is different since we do not have a priori which is the relation between $\alpha$ and $N$.

 \subsection*{Algorithm to find $\alpha N$  and N}

%We start with  $\alpha_0=\frac{\max(S)}{\mbox{Coef} \tilde{I}(0)}$ and 
%${T}=Sn.$
 We start with an initial  $\alpha_0$ and ${T}=S_{dev}.$
\medskip
At each step $i$,
\begin{enumerate}
\item Calculate $\max {T}$ and $b_k$ the point where  it attains the maximum.
\item Redefine
$$T(x)=T(x)-\alpha_0 \sum_{k=1}^{i} \tilde{I}_{dev}(x-b_k)$$
\item
$t(k)=\|T\|
$
\item  The algorithm stops when $t$ arrives to a minimum, and the number of sources used when that minimum is reached is the selected value for $N$.
\end{enumerate}

At the end $\alpha_0 N$ approximates $\sumi R(x_i)$.

Now this is our $N$. The method now follows as in the previous subsection, the only difference is that at the end of step 2) we add a step:

2b) Once we have found $\{\tilde{a_k}\}_{k=1}^N$ we use a linear Least Squares fit to find a corrected value for $\alpha$. 

$\alpha N$ is an invariant that also approximates $\sumi R(x_i)$.

\subsection*{Notation}

To contemplate both cases we denote $S_*=S$ and $\tilde{I}_*(x)  =\tilde{I}$ when there is no background and $S_*=S_{dev}$ and $\tilde{I}_*(x)  =\tilde{I}_{dev}$ when we are in the case with background.

We use the following notation:
\begin{align*}
&d_0=2 \times \mbox{ standard deviation of } I(x),\ 
d_p=\mbox{pixel size},\\ 
&d_s=\mbox{superpixel, } 
\end{align*}

When $R$ is discrete we can denote  $\{y_p\}_{p=1}^m$ the points where  $R$ is supported, being  $m$ the total number of such points. In this case $R_p$ is the intensity of $R$ on each $y_p$. 

 Therefore,
\begin{align}\label{defR} R(x)= \sum_{p=1}^m R_p \delta(x-y_p).\end{align}
If $R$ is continuous we are going to assume that $R$ can by approximated by  \eqref{defR}. For the propose of this paper we assume $R$ is discrete. We are not going to give a bound of the error due to this discrete  approximation. In general the number $m$ where $R$ is supported  is assumed to be large.  
%can approximate it using the superpixel $d_s$ and defining for this case:
%$$R_p=R(y_p)d_s^D.$$

We approximate $R$ by $\bar{R}$ as a superposition of point sources of identical intensities $\alpha$  
and denote their positions as $\{a_k\}_{k=1}^N$ (taking into account the repetitions). 
We define \begin{align*} N_p:=\left[\frac{R_p}{\alpha}\right]\quad \mbox{ and }\quad 
\bar{R}_p:=N_p \alpha.\end{align*}

If $\sump N_p<N$ we order  $\frac{R_p}{\alpha}-[\frac{R_p}{\alpha}]$ 
decreasingly and we add $1$ at each $N_p$ until 
$\sump N_p=N$ and we redefine, 
$$\bar{R}_p:=N_p \alpha.$$
Therefore
\begin{align*} a_k=y_1,& \mbox{ with } k=1,..,N_1,\\ 
      a_k=y_p, & \mbox{ with }     k=\sum_{r=1}^{p-1}N_r+i-1:\sum_{r=1}^{p-1}N_r+N_p+p-1.
\end{align*}
On the other hand,  $\bar{R}_p=\alpha\ \sharp\{k: a_k=y_p\}.$ Therefore we have,
\begin{equation}\label{Rptecho}
\alpha \sum_{k=1}^N I(x_i-a_k) =\sum_{p=1}^m \bar{R}_p I(x-y_p).
\end{equation}
This identity is only used to obtain the bounds for the uncertainties. Remember that  since for each  $k=1,\cdots,N$, $a_k \in \R^D$   our space for optimization is $\R^N\times \R^D$ for a fixed $\alpha$.

We denote $\| \cdot\|$ to the standard $2-$norm. Depending on the context will be taking the norm in $\R^n$ or in $\R^D$.

We denote $\langle \cdot,  \cdot \rangle $ as an average over the ensemble of possible realizations of the measurement, not an average over many actual measurements. 

To have a complete list of all the variable used see Table \ref{tab:def}.
% needed in second column of first page if using \IEEEpubid
%\IEEEpubidadjcol

\section{Uncertainties of the reconstruction and optimum value for $N$}
\label{sec.errores}
%%%%%%%%%%%%%%%%%%%%%%%%%%%%%%%%%%%%%%%%%%%%%%%%%%%%%%%%%%%%%%%%%%%%%%%
The quality of the reconstruction depends on the number of sources $N$ used.  
 To find the best choice for $N$ we define the uncertainty in the positions 
  $\sigma$ as,
$$\sigma^2=\frac{1}{N}\min_{\tau}\sum_{k=1}^N \langle\|{\delta_k}^{\tau}\|^2\rangle=\frac{1}{N}\sum_{k=1}^N \langle\|{\delta_k}^{\tau_0}\|^2\rangle$$
where ${\delta_k}^{\tau}=a_{\tau(k)}-\tilde{a}_k$ and  $\tau$ is any permutation of the set $\{1,\cdots,N\}$.  To simplify the notation along the paper we will omit  the letter $\tau_0$ assuming  that $a_k$ has the correct order.
We will find a bound for $\sigma$ that will depend on $N$ and finally determine the value for $N$ that minimizes that bound that we call $N_{op}$   (see \eqref{Nopinal}). For this purpose we will need to find a bound of the  error due to the fit of $I$ (see \eqref{cotaItilde}) and on the error due to the truncation on $R$ (see \eqref{cotaRbar}).

 We will define the super-resolution factor $M_s$ as 
\begin{equation}\displaystyle{\label{Ms}
\begin{aligned}
M_s=\frac{d_0}{2\sigma}.
\end{aligned}}
\end{equation}

Along the forthcoming calculations an error of lower order will arise when computing functions of the positions shifted by a fraction of the size of the pixel. That is, given $a\in \R^D$ and any derivable function $f:\R \to \R$  we have
$$\sum_{i=1}^n f(x_i-a)=\sum_{i=1}^n f(x_i)+E_a$$
and we want to estimate  $E_a$. 

Suppose that $f$ and  $f(x-a)$ have support in the same region. Recall that one usually fits a small portion of the complete data set ignoring what happens at the boundaries.     
We can define for each  $a$,
$P(a)$ the nearest pixel to  $a$, then
$$\sum_{i=1}^n f(x_i-a)=\sumi f(x_i-a+P(a))$$
making a first order approximation and using Cauchy- Schwartz inequality we have,
\begin{equation}\displaystyle{\label{errorsuma}
\begin{aligned}
&E_a=|\sum_{i=1}^n f(x_i-a)-\sum_{i=1}^n f(x_i)|
\sim |\sum_{i=1}^n \nabla f(x_i)(a-P(a))|\\
& = |\sum_{j=1}^D \sum_{i=1}^n  \frac{\partial f}{\partial x_j}(x_i)(a^j-P(a^j))|\\
& \leq  \| \sum_{i=1}^n  \nabla f(x_i)\| \| a-P(a)\| 
 \leq  \sqrt{2}^{d-1}\left(\frac{d_p}{2}\right) \| \sum_{i=1}^n  \nabla f(x_i)\|.
\end{aligned}}
\end{equation}

\begin{remark}\label{Edespreciable}
In the cases that   $f=h^2(x)$ and $h(x)$  is even or odd in all the coordinates, then the partial derivatives are odd or even (respectively) therefore the sum
$$\sumi h(x_i) h_{x_j}(x_i)= 0$$
then the term of first order in the sum is zero and we can suppose that  $E_a$ is 
negligible.  Also observe that in the case that  the pixel size is small this terms are also negligible.
\end{remark}
\subsection{Steps of the estimation}
Recall that  $\bar{R}$ is the truncation of $R$ and  $\tilde{R}$ where the minimum is attained. Then we have that 
\begin{equation}\label{cotaminima}\|S_* - \tilde{R}*\tilde{I}_*\|\leq \|S_* - \bar{R}*\tilde{I}_*\|\end{equation}

We alse are going to use the following inequality that is a direct consecuence of Young's inequlity. We have that for any $\varepsilon >0$,  \begin{equation}\label{lapapa}a^2+b^2+2 ab\leq a^2(1+\varepsilon)+b^2(1+1/\varepsilon).\end{equation}
%On the other hand, when there is no background we can use that
%\begin{align*} S(x) - \bar{R}*\tilde{I}(x)&=
%{R}*{I}(x)- {R}*\tilde{I}(x)\\&+{R}*\tilde{I}(x)-\bar{R}*\tilde{I}(x)+\eta(x).\end{align*}
In the following we are going to
take the average  over all the possible realizations of $S$ (i.e. of eventual different measurements of the same event),  that means  that  $S$ is a random variable and that this randomness depends only on the noise. 

Therefore adding and subtracting the term ${R}*\tilde{I}(x),$   calling 
 $U=({R}-{\bar{R}})*\tilde{I}$ and $V={R}*({I}- \tilde{I})$ we have,
\begin{align*}\label{cotaminima2} \langle \|S -& \bar{R}*\tilde{I}\|^2  \rangle \\&=
    \langle \|V+ \eta\|^2 + \|U\|^2 \rangle
 +2 \langle \sum_{i=1}^n  U(x_i)(V(x_i) + \eta(x_i)
 \rangle\\&=   \langle \|V \|^2 +\| \eta\|^2 + \|U\|^2 \rangle+2 \langle \sum_{i=1}^n U(x_i)(V(x_i) \\&+ \eta(x_i))\rangle+2\langle \sum_{i=1}^n V(x_i) \eta(x_i))\rangle.
  \end{align*}
 Using that for each $i$, $\langle \eta(x_i) \rangle=0$ and since the average only depends on the noise, we have that,
$$ \langle \sum_{i=1}^n  U(x_i) V(x_i) + \eta(x_i)
 + V(x_i) \eta(x_i)\rangle=   \sum_{i=1}^n U(x_i) V(x_i). 
$$
Using inequality \eqref{lapapa} we have that for any $\varepsilon >0$
\begin{equation}\label{cotaminima3} \langle \|S-\bar{R}*\tilde{I}\|\rangle \leq (1+1/\varepsilon)   \|V \|^2 + (1+\varepsilon)\|U\|^2 + \langle \| \eta\|^2 \rangle
\end{equation}

In the case we have an unknown background that we assume constant, we can use that that 
 $\langle  B-\frac{1}{n}\sum_{i=1}^n B(x_i)\rangle =0$ and that $\sum_{i=1}^n \eta(x_i)  =0$ (see the discussion in Section\ref{background}) therefore,  
 \begin{equation}\label{cotaminimafondo} \langle \|S_*-\bar{R}*\tilde{I}_{dev}\|\rangle \leq (1+1/\varepsilon)   \|V \|^2 + (1+\varepsilon)\|U\|^2 + \langle \| \eta\|^2 \rangle
\end{equation} 
 where here $U=({R}-{\bar{R}})*\tilde{I}_{dev}$.

In conclusion, to estimate the quadratic error we have to estimate three terms:
\subsubsection{Error due to the noise}: That is the last term in \eqref{cotaminima3} and \eqref{cotaminimafondo}.
\subsubsection{ Error due to the fit of $I$}
We are assuming that we have an approximation $\tilde{I}$ of $I$ and we want to estimate that term. We also are assuming that all the functions have support in a ball of radius  $d_0$ or that that the functions decrease very fast  when $|z|> d_0$.
In that case, we obtain,
\begin{equation}\label{cotaItilde}
\begin{aligned}
 \sum_{i=1}^n&({R}*{I}(x_i)- {R}*\tilde{I}(x_i))^2\\&= \sumi \left ( \sump R_p (I(x_i-y_p)-\tilde{I}(x_i-y_p))\right)^2 \\&
= \sump R_p^2 \sumi( g(x_i-y_p))^2\\&+ 2  \sump \sum_{l\neq p} R_l R_p \sumi  g(x_i-y_p) g(x_i-y_l)\\
&=\sump \left(R_p^2G(0)+2 \sum_{l\neq p : \|y_p-y_l\|< d_0}  R_l R_p G(y_l-y_p)\right)\\ &\quad +\underbrace{\sump \left(R_p^2 E_{y_p}+ 2   \sum_{l\neq p : \|y_p-y_l\|< d_0} R_l R_p  E^l_{y_p} \right)}_{E_G}:=A
\end{aligned}\end{equation}
Where,
$
g(x)=\tilde{I}(x)-I(x),$ 
$$G(z)=\sumi g(x_i) g(x_i-z),
$$
and 
\begin{equation}\label{Gylyp} \sumi g(x_i-y_p) g(x_i-y_l)=G(y_p-y_l)+E^l_{y_p}.\end{equation}
For $E^l_{y_p}$ we have the estimate \eqref{errorsuma} for the function  $f(x)=g(x) g(x-y_l+y_p)$ when $l\neq p$ and for  $E_{y_p}$ the estimate  $f(x)=g^2(x)$. 

In the practical examples we will drop the term  $E_G$ (which is of lower order  because in all practical applications of the method the measurements must be oversampled).

%To simplify the estimate we can apply the same argument as before and obtain, 
%$$\sumi g^2(x_i-(y_p-y_l))= \sumi g^2(x_i) +E_{y_p- y_l}$$
%having  $E_{y_p- y_l}$ the estimation  \eqref{errorsuma} for the function $f(x)=g^2(x)$. 
%
%Therefore,
%\begin{align*}
%G&(y_p-y_l)
%\\&=\sumi g(x_i) g(x_i-(y_l-y_p))\\ &\leq (\sumi g^2(x_i))^{1/2} (\sumi g^2(x_i-(y_l-y_p))^{1/2} \\ &=
% (\sumi g^2(x_i))^{1/2} \left\{\sumi g^2(x_i)+E_{y_p- y_l})^{1/2}\right\}
%\\ & \sim (\sumi g^2(x_i))^{1/2}  \left\{(\sumi g^2(x_i))^{1/2}+\frac{E_{y_p- y_l}}{2 (\sumi g^2(x_i))^{1/2} }\right\}\\
%&= \sumi g^2(x_i)+ \frac{E_{y_p- y_l}}{2}=G(0)+ \frac{E_{y_p- y_l}}{2},
%\end{align*}
%then
%\begin{align}\label{cotaA}
%A&\leq \left(\sump R_p^2 + 2\sump R_p \sum_{l\neq p : \|y_p-y_l\|< d_0} R_l \right) G(0)+E_G
%%\\
%%&=:F
%%\\
%%&\leq \left(\sump R_p^2 + 4M_s m (max R)^2\right) G(0)+ E_G
%\end{align}
%and 
% $E_G$ is a global bound of the error. Observe that in the second term only appears $R_p$ multiplied by the neighbours, when this source is isolated this term does not appear. 
%
%To eliminate  $E_G$ we use that the following term is negligible
%\begin{align*}\label{Cg}C_{g_{lp}}=\|\sum_{i=1}^n \nabla g(x_i)g(x_i-y_l+y_p)\\+g(x_i)\nabla  g(x_i-y_l+y_p)+\nabla g(x_i) g(x_i)\|.\end{align*}
\subsubsection{ Error due to the truncation on $R$} 
%\item {\bf Error debido al truncado de $R$.}
\medskip
For the truncation error we have, calling $X_p=\bar{R}_p-R_p$,
\begin{align*}&\sum_{i=1}^n(({R}-\bar{R})*\tilde{I}_*(x_i))^2\\
&=\sum_{i=1}^n(\sum_{p=1}^m{X_p}\tilde{I}_*(x_i-y_p))^2=\sum_{i=1}^n\sum_{p=1}^m{X_p^2}(\tilde{I}_*(x_i-y_p))^2 \\&+2 \underbrace{\sum_{i=1}^n\sum_{p=1,l\neq p}^m{X_p X_l}\tilde{I}_*(x_i-y_p) \tilde{I}_*(x_i-y_l)}_{Mix}.
\end{align*}
Since each  $\bar{R}_p-R_p$ belongs to  $[-\alpha/2, \alpha/2]$  and there is no correlation between the value of $\alpha$ and the hight of the spike $R_p$ we can think that for $m>>1$,  $X_p$ follows a uniform distribution in  $[-\alpha/2, \alpha/2]$, and,
\begin{equation}
\label{estimuniforme}
\frac{1}{m}\sum_{p=1}^m(\bar{R}_p-R_p)^2\sim\alpha^2/12.\end{equation}
 If $m$ is small  we can  replace $\alpha^2/12$ by $\alpha^2/4$.  

On the other hand, if we denote 
\begin{align}\label{defY}
Y(z)&=\sumi |\tilde{I}_*(x_i)| |\tilde{I}_*(x_i-z)|
\end{align}
using an argument similar to  \eqref{cotaItilde} we arrive at,
\begin{align*}
 Mix=&\sum_{i=1}^n\sump \sum_{l\neq p}{X_p X_l}\tilde{I}_*(x_i-y_p) \tilde{I}_*(x_i-y_l)\\
\leq&\frac{\alpha^2}{4}\left( \sump \sum_{\{l\neq p:\|y_p-y_l\|<d_0\}} Y(y_l-y_p)\right.\\ &\left.+ \sump\sum_{\{l\neq p:\|y_p-y_l\|<d_0\}} E^l_{y_p}\right).
\end{align*}
%\begin{align*}
% Mix=&\sum_{i=1}^n\sum_{p=1,r\neq p}^m{X_p X_l}\tilde{I}_*(x_i-y_p) \tilde{I}_*(x_i-y_l)\\
% =&\sum_{i=1}^n\sump \sum_{l\neq p}{X_p X_l}\tilde{I}_*(x_i-y_p) \tilde{I}_*(x_i-y_l)\\
%\leq&\frac{\alpha^2}{4}\left( \sump \sum_{\{l\neq p:\|y_p-y_l\|<d_0\}} Y(y_l-y_p)\right.\\ &\left.+ \sump\sum_{\{l\neq p:\|y_p-y_l\|<d_0\}} E^l_{y_p}\right).
%\end{align*}
% \begin{align*}
%\sum_{i=1}^n(({R}-\bar{R})*\tilde{I}_*(x_i))^2\sim &\frac{\alpha^2}{12} \left( m \sum_{i=1}^n(\tilde{I}_*(x_i))^2 + \sump E_{y_p}\right).
%\end{align*}
For $E^l_{y_p}$ we have the estimate \eqref{errorsuma} for the function $f(x)=|\tilde{I}_*(x)| |\tilde{I}_*(x-y_l+y_p)|$ when  $l\neq p$ and for  $E_{y_p}$ the estimate with the function $f(x)=\tilde{I}_*^2(x)$. 

Combining this with  \eqref{estimuniforme} we finally 
obtain the bound due to the truncation,
\begin{equation}\label{cotaRbar}
\begin{aligned}
\sum_{i=1}^n(({R}-\bar{R})&*\tilde{I}_*(x_i))^2\lesssim \frac{\alpha^2}{12} \left(m\sum_{i=1}^n(\tilde{I}_*(x_i))^2 \right.\\&\left.+ 3  \sump \sum_{\{l\neq p:\|y_p-y_l\|<d_0\}} Y(y_l-y_p)+E_{\bar{R}}\right),
\end{aligned}
\end{equation}
where
\begin{align*}E_{\bar{R}}\leq & \sqrt{2}^{d-1}d_p \left(m\| \sum_{i=1}^n  \nabla \tilde{I}_* (x_i) \tilde{I}(x_i)\|\right.\\&+\left.\sump \sum_{l\neq p}\| \sum_{i=1}^n  \nabla \tilde{I}_* (x_i-y_l+y_p) \tilde{I}(x_i-y_l+y_p) \right.\\ &\left. +\nabla \tilde{I}_* (x_i-y_l+y_p) \tilde{I}(x_i-y_l+y_p)\|\right).
\end{align*} 
In the cases that we have parity (see
 Remark \ref{Edespreciable}) we can drop the term   $E_{\bar{R}}$.

\medskip
\begin{remark}
In the cases that the $y_p$ are all isolated   at distance more than  $d_0$ the term  $Mix$ will be zero. In other cases the function  $Y(y_l-y_p)$ decreases as the sources separate.
\end{remark}

\medskip

\subsubsection{Errors due to $\delta_k$} 
%\item {\bf Error en la posici\'on }
%Dropping the lower order terms we have,
%\begin{align*}
%\alpha^2 \langle\sum_{i=1}^n& (\sumk(\tilde{I}(x_i-a_k)-\tilde{I}(x_i-\tilde{a_k}))^2\rangle\\
% \sim  &
%\alpha^2 \sum_{i=1}^n \sum_{k=1}^N\langle(\nabla \tilde{I}(x_i-a_k)\cdot\delta_k)^2 \rangle
%\end{align*}
If we take the average over all the possible realizations  we have,
\begin{align*}
\langle(\sum_{k=1}^N &\nabla \tilde{I}(x_i-a_k)\cdot\delta_k)^2\rangle\\&=  \sum_{k=1}^N \left\{\sum_{j=1}^D \left(\frac{ \partial\tilde{I}}{\partial x_j}(x_i-a_k)\right)^2\langle (\delta_k ^j)^2 \rangle\right.\\&\left. + 2 \sum_{s\neq j}^D \frac{ \partial\tilde{I}}{\partial x_j}(x_i-a_k) \frac{ \partial\tilde{I}}{\partial x_s}(x_i-a_k) \langle \delta_k ^j \delta_k ^s \rangle \right\}\\&+
2 \sum_{l\neq k}^N \sum_{j=1}^D\sum_{s\neq j}^D \frac{ \partial\tilde{I}}{\partial x_j}(x_i-a_k) \frac{ \partial\tilde{I}}{\partial x_s}(x_i-a_l)  \langle \delta_k ^j \delta_l ^j \rangle 
\end{align*}
Dropping the lower order terms  and using that  $ \delta_k $ and $\delta_l$ are independent as well as  $\delta^{i}_k$ and $\delta^{j}_k$,
we have
\begin{align*}
 \langle\sum_{i=1}^n &\left(\sumk(\tilde{I}(x_i-a_k)-\tilde{I}(x_i-\tilde{a_k})\right)^2\rangle
\\ \sim &   \sum_{i=1}^n \sum_{k=1}^N \sum_{j=1}^D \left(\frac{ \partial\tilde{I}}{\partial x_j}(x_i-a_k)\right)^2 \langle (\delta_k ^j)^2 \rangle
\\\geq&  \left(\sum_{k=1}^N    {\min_{s=1,...,D}}\sum_{i=1}^n\left(\frac{ \partial\tilde{I}}{\partial x_s}(x_i)\right)^2 \langle\ \|\delta_k\|^2\rangle \right. \\&\left.+\sum_{j=1}^D E_{a_k}^j \langle (\delta_k ^j)^2 \rangle \right)\\\geq
&  N^2 \frac{\sigma^2}{N} \left( 
 {\min_{s=1,...,D}}\sum_{i=1}^n\left(\frac{ \partial\tilde{I}}{\partial x_s}(x_i)\right)^2  -E_{\sigma} 
\right)\end{align*}
where in the last inequality  $$E_{\sigma}=\sqrt{2}^{d-1}\left(\frac{d_p}{2}\right){\max_{t=1,...,d}}\left\|\sum_{i=1}^n\nabla \left(\frac{ \partial\tilde{I}}{\partial x_t}(x_i)\right)^2\right\| 
$$ and we are using the estimate  of the error for the translation of the sum where $E_{a_k}^j$  is estimated by  \eqref{errorsuma} taking $f(x)= (\tilde{I}_{x_j})^2(x)$.
In this case we have,
$$C:={\min_{s=1,...,D}}\sum_{i=1}^n\left(\frac{ \partial\tilde{I}}{\partial x_s}(x_i)\right)^2  -E_{\sigma} >0$$
%When we have parity  of 
When $\tilde{I}$ has a well defined parity, then by  Remark \ref{Edespreciable}  the term  $E_{\sigma}$ is 
negligible.

If  we define 
\begin{equation}\label{defkappa2}\kappa''=\alpha N\sqrt{C}
\end{equation}
we have,
\begin{equation}\label{kappa2}(\kappa'')^2 \frac{\sigma^2}{N}\leq \alpha^2 \langle\sum_{i=1}^n (\sumk(\tilde{I}(x_i-a_k)-\tilde{I}(x_i-\tilde{a_k}))^2\rangle.\end{equation}

%\end{enumerate}

\subsection{Estimation of the optimal  $\sigma$}
Using  \eqref{cotaminima},  \eqref{cotaminima3} and \eqref{cotaminimafondo},
\begin{align*}\alpha^2&\langle \sum_{i=1}^n (\sum_{k=1}^N(\tilde{I}(x_i-a_k)-\tilde{I}(x_i-\tilde{a_k}))^2\rangle\\ &= \langle \|(\langle\tilde{R}-\bar{R})*\tilde{I}_*\|^2\rangle \\
&\leq \langle (\|\tilde{R}*\tilde{I}_*-S_*\|+\|\bar{R}*\tilde{I}_*-S_*\|)^2\rangle\\
&\leq  4 \langle\|\bar{R}*\tilde{I}_*-S_*\|^2\rangle\\
&\leq  4((1+1/\varepsilon)   \|V \|^2 + (1+\varepsilon)\|U\|^2 + \langle \| \eta\|^2 \rangle).
\end{align*}
Denoting,
\begin{equation}\label{defF} F= \sump \{R_p^2G(0)+2\sum_{l\neq p : \|y_p-y_l\|< d_0}  R_l R_p G(y_l-y_p)\}\end{equation}
\begin{equation}\label{kappa}\kappa^2:=4( (F+ E_G) (1+1/\varepsilon)+\langle\|\eta\|^2\rangle)\end{equation}
and
\begin{equation}\label{defL}
L=\|\tilde{I}_*\|^2 + \frac{3}{m}  \sump \sum_{l\neq p : \|y_p-y_l\|< d_0} Y(y_l-y_p)
\end{equation}

\begin{equation}\label{kappa1}
\begin{aligned}(\kappa')^2= \frac{4(1+\varepsilon) \alpha^2 N^2 m}{12}  & (L+E_{\bar{R}})\end{aligned}\end{equation}

%$$ \left(\|\tilde{I}_*\|^2 + \frac{3}{m} \sump \sharp\{ l\neq p:\|y_p-y_l\|<d_0\} Y(y_l-y_p)+E_{\bar{R}}\right)$$
We have by \eqref{cotaItilde},  \eqref{cotaRbar} and \eqref{kappa2}  that
%$$(\kappa'')^2 \frac{\sigma^2}{N}\leq \frac{(\kappa')^2}{N^2}+\kappa^2$$
%o sea
\begin{equation}\label{cotasigma}\sigma^2\leq \frac{\kappa'^2}{(\kappa'')^2 N}+\frac{\kappa^2}{(\kappa'')^2} {N}.\end{equation}

Since we are assuming that $\alpha N$ is  an invariant, which approximates the sum of $R$, we will denote  $\alpha N=Z$ (See subsection {\it Algorithm to find $\alpha N$}).

Then we have an estimate of the optimal  $N$ ($N_{op}$) where the right hand side of \eqref{cotasigma} is minimum, and a $\sigma=\sigma_{op}$ that is when
\begin{align}\label{Nop}
\begin{split}
N_{op}&=\frac{\kappa'}{\kappa}\quad \mbox{ and }\\
\sigma_{op}^2&=  \frac{2\kappa'\kappa}{(\kappa'')^2}.
\end{split}
\end{align}

Therefore,   in the case that  the pixel size is small,  we can drop  the terms corresponding to the errors due to the translations and therefore,
\begin{align*}\displaystyle \sigma^2 \leq &\frac{4 (1+\varepsilon) m  L}{ 12N \tilde{I}_{der}^2  } +  \frac{4N((1+1/\varepsilon)F+\langle\|\eta\|^2\rangle)}{Z^2 \tilde{I}_{der}^2 } . \end{align*} 
where
\begin{equation}\tilde{I}_{der}= {\min_{s=1,...,D}}\left \|\frac{\partial\tilde{I}}{\partial x_s}\right\|. \end{equation}
And we also can see that this function attains a minimum when 
\begin{align}\label{Nopinal} N_{op}&= \frac{ (1+\varepsilon)^{1/2} m^{1/2}}{\sqrt{12}}   {\left( \frac{Z^2 }{(1+1/\varepsilon){F}+\langle\|\eta\|^2\rangle}\right)^{1/2}} L^{1/2} \end{align} 
and at
\begin{align}\sigma_{op}&= \frac{2 (1+\varepsilon)^{1/4}  m^{1/4} }{{3}^{1/4} \tilde{I}_{der}}  \left( \frac{{F}(1+1/\varepsilon)+\langle\|\eta\|^2\rangle}{Z^2 }\right)^{1/4}L^{1/4} .\end{align}
If the number of actual sources $m$ is small, after replacing $\alpha^2/12$  by $\alpha^2/4$ in \eqref{estimuniforme} we obtain the same $\sigma_{op}$ but without the factor ${3}^{1/4}$.

%Observe that $N_{op}the definition of  \eqref{defF}  

Notice that in the first parentheses the numerator is constant and the denominator has two contributions, one from the error in the determination of $I$ and the other one arises from the noise in the measurement, decreasing any of them increases the optimum $N$. On the other hand if the sources are isolated the term with  $Y(y_l-y_p)$ in $L$ and $G(y_l-y_p)$ in $F$ disappear. Hence sparsity of the source helps the quality of the reconstruction, although as it will be shown in the examples this is not a requirement as strict as with compressed sensing schemes because once the source is sparse enough the other terms dominate. Also notice that $F$ can be reduced by improving the measurement and fit of $I$ remaining only the term from the noise.
\begin{remark}
Observe that the optimal $N$ ($N_{op}$) cannot be calculated a priori.  That is why in practical applications, we have to start first with an arbitrary $N$, and then for that $N$, find an intermediate minimum to be able to compute all the terms needed to calculate $N_{op}$.
\end{remark}

\subsection{Bound for $\chi^2$}
We can also determine a bound for $
\chi^2$. In fact again by, \eqref{cotaminima},  \eqref{cotaminima3},\eqref{cotaminimafondo} \eqref{cotaItilde},  \eqref{cotaRbar}  and neglecting again the terms involving errors due to translations ($E$) we have for any $\varepsilon >0$
%\begin{align*}
%&\langle\|S - \tilde{R}*\tilde{I}\|^2\rangle \leq\\&  (1+\frac{1}{\varepsilon})\left(\sump \{R_p^2G(0)   +2\sum_{l\neq p : \|y_p-y_l\|< d_0}  R_l R_p G(y_l-y_p)\} \right)\\  & + (1+\frac{1}{\varepsilon})E_G+ \langle\|\eta\|^2\rangle\\
%&+\frac{ (1+\varepsilon) \alpha^2  m}{12}  \left(\|\tilde{I}_*\|^2 + \frac{3}{m} \sump \sum_{\{l\neq p:\|y_p-y_l\|<d_0\}} Y(y_l-y_p)\right)\\ & +\frac{(1+\varepsilon) \alpha^2  m}{12}E_{\bar{R}}. 
%\end{align*}
\begin{align*}
&\langle\|S_* - \tilde{R}*\tilde{I}_*\|^2\rangle  \leq\\ 
&  (1+\frac{1}{\varepsilon}){\left(\sump R_p^2G(0)+2\sum_{l\in U_p}  R_l R_p G(y_l-y_p) \right)} 
+ \langle\|\eta\|^2\rangle\\
&+\frac{(1+\varepsilon) \alpha^2  m}{12}  \left(\|\tilde{I}_*\|^2 + \frac{3}{m} \sump \sum_{l\in U_{p}} Y(y_l-y_p) \right),\end{align*}
where $U_p=\{l\neq p:\|y_p-y_l\|<d_0\}.$
\subsection{Resolving faint sources vs. $\sigma$} 
 \medskip 
 
The value obtained for $N_{op}$ is the one that optimizes the resolution, but it could be at the expense of not resolving faint sources (due to the discrete resolution in intensity given by $\alpha$). Hence a better resolution in height might require to pay a price in $\sigma$, particularly important to make visible points that faded away. 

To analyze this compromise lets look at the relation between them. From \eqref{cotasigma}, and \eqref{Nop} it results
\begin{equation}\label{sigmaN}
\frac{\sigma^2}{\sigma_{op}^2}=\frac{1}{2}\left(\frac{N_{op}}{N}+\frac{N}{N_{op}}\right)\end{equation}

%If $$\sigma^2= \frac{s}{N}+t N$$
%then as in  \eqref{Nop} we have that 
%$N_{op}=\frac{\sqrt{s}}{\sqrt{t}}$ and
%$\sigma_{op}^2=2 N_{op}t.$
 
% Hence making  $r=\frac{\sigma}{ \sigma_{op}}$ and $u=\frac{N}{N_{op}}$ we get
%\begin{equation}\label{sigmaN}
%r^2=\frac{\sigma^2}{\sigma_{op}^2}= \frac{\frac{s}{N}+t N}{2N_{op}t}=
%\frac{1}{2}\left(\frac{N_{op}}{N}+\frac{N}{N_{op}}\right)=
%\frac{1}{2}\left(u+\frac{1}{u}\right)\end{equation}

This gives a straightforward relation between the increase in $N$ and the increase in $\sigma$. Doubling the sensitivity in height (double $N$) only degrades the resolution by $11\%$ and doubling the resolution $\sigma$ increases the height sensitivity by an order of magnitude. Hence the optimum value obtained before in practice might not be the best compromise, depending on the information we are seeking from the measurement.

%Como nosotros vamos a tener una cota de $\sigma$ nos va a decir en realidad que un %aumento del $N$ va a dar que el factor $r$ caiga por debajo de esta curva.

 \section{Examples}
\label{sec.resultados}

\begin{exmp}
With known background ($D=1$).

For an example in one dimension and with a known and substractable background we chose the deconvolution of spectral lines acquired with an array spectrometer. The instrument chosen was a Thorlabs CCS200/M that has a nominal spectral resolution of 1nm and a pixel of 0.22nm. This provides an adequate oversampling for the method. The instrument was used to measure spectral lines from pure gas discharge lamps in order to use isolated lines to determine the IRF and overlapping lines to show the power and limitations of the method. The background was measured and subtracted by taking for each measurement a spectrum with the lamp off. Hence for this example we have $D=1$ and $B=0$. 
For each measurement after subtracting the background the signal was divided by the spectrum collected from an incandescent lamp. In this manner pixel to pixel differences in the detector array are corrected, and assuming the incandescent lamp spectrum is constant within the spectral line to be resolved (deconvolved) no significant distortion from this normalization arises. 
For the determination of the noise we measured the signal from an incandescent lamp at least 100 times for several intensities (obtained by changing the distance from the lamp to the input optical fiber of the spectrometer), and for each pixel and each intensity (counts) the standard deviation is calculated.

If $S_{lamp}$ is the measure of the incandescent lamp, $S$ of the discharge lamp then measuring   the dark background $B$ (with the source off) we can define new variable $S_{norm}=\frac{S-B}{S_{lamp}-B}$.
As mentioned the denominator is a smooth function and can be considered as a constant along the zone we want to solve. We replace  then the original spectra by these new background free re-normalized one.

%The function selected for the fit was a result of several trials with asymmetric functions until an adequate bound for g was achieved. In this manner $g$ resulted  small enough to make its contribution to the uncertainties negligible when compared to those arising from the noise of the measurement in \eqref{kappa}.

The determination of a fit function for the IRF is described in Apendix \ref{espectro}.

To test the method for this example we take from the NIST table a double peaks of Na, one at $589.00$nm of intensity $1000$ and the other one at $589.59$nm with intensity $500$.
%\begin{align*}
%Na &\ 1000 &  589 \\ 
%Na &\ 500    & 589.59 
%\end{align*}
The results obtained are shown in Figure \ref{fig:Na}.

In this case we start with $N=100$ after using the algorithm  the  $N_{op}$ obtained using our bounds is $5$ and the $\sigma_{op}=0.33$nm when $d_a=0.22$nm and $\sigma_{op}=0.39$nm with $d_a=0.11$nm. 

On the other hand, compared with the NIST table,  when we look at the histogram with $d_a=0.11$nm  the error is lower than $2\sigma=0.22$nm, showing that our estimated bound is a factor of 3 larger than  the actual uncertainty. For this fit the  super-resolution factor (improvement in the resolution)  is $M_s> 5.6$ and the predicted improvement was   $\frac{d_0}{2\sigma_{op}}= 1.85.$ %Moreover when $d_a=0.22$ nm we obtained that $m_a=3$ instead of $m$. That introduces a factor $1.5^{1/4}=1.12$ in the term $\sigma_{op}$ so in this case we can say that $\sigma_{op}$ is less than $0.33/1.122$nm$=0.29$nm very close to the empirical observation. In conclusion if we have the additional information of the number of peaks this number $m_a$ can be replaced by $m$.

\begin{figure}
    \centering
  \includegraphics[width=3in]{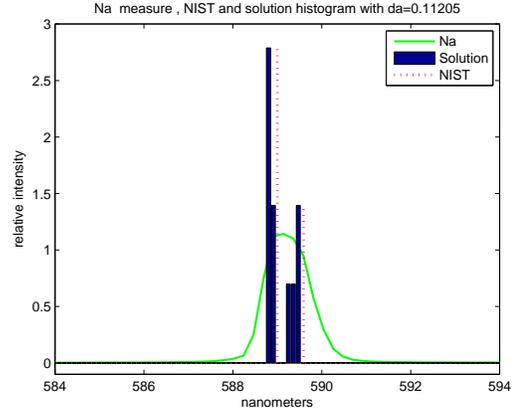}
    \caption{Results and comparison with NIST for the 2 peaks of Na. In the y axis we are representing the relative intensity}\label{fig:Na}
\end{figure} 
Then we take a zone of the spectrum of the $Kr$ where there is a intense peak at $557.03$nm of intensity $300$ and in both sides two peaks  of lower intensity one at $556.22$nm with intensity $80$ and the other at $558.04$nm with intensity $13$.   
%\begin{align*}
% Kr\\&
%   80\ &	556.22\\
%   & 300    & 557.03\\
%  & 13\ &	 558.04
%\end{align*}
Here besides the closeness of the peaks we have to deal with the large difference in the intensities.
% the problem is not the 
%closeness of the peaks but  the difference of %intensity.

So we take $N=100$, we are not taking the optimal $N$ because the optimal will not find the lower peaks.
After using our method and using the calibration we obtained the  results  of Figure \ref{fig:Kr}.

We can see that the method can distinguish one of the two peaks of low intensity, but the lowest is masked by the noise and could not be recovered.

This example is sparse enough that it could also be deconvolved using compressed sensing. We used it to show a practical example where the ground truth is known to test the predictions.

  \begin{figure}
    \centering
 \includegraphics[width=3in]{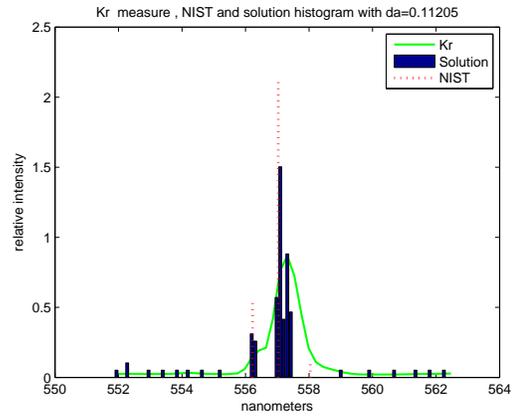}
    \caption{Results and comparison with NIST for the 3 peaks of Kr. In the y axis we are representing the relative intensity}\label{fig:Kr}
\end{figure} 

 \end{exmp}
 \medskip

\begin{exmp} Synthetic Image with unknown background.

\medskip
As for the real fluorescent images discussed in the next example we do not have a ground truth to validate the deconvolution, we started with an artificially generated image before going to real situations. For this purpose the pixel size and resolution of our experimental data were used, i.e. a pixel size of 68nm, a dynamic range for the camera of 16 bits and a noise figure for an image $S$ of a standard deviation of $23+\sqrt{S}$. The point spread function (IRF) of the microscope was assumed Gaussian with a standard deviation =1.435 pixels=97.6nm. This is similar to the resolution obtained for 520nm light with an objective with a numerical aperture of 1.3 and yields a resolution of 200nm defined as the distance between two point sources of equal intensity at which the two maxima start to be visible (assuming no noise). The test image was  synthesized by convolving the IRF with a source $R$ determined by two parallel straight segments 144nm apart, each generated by 71sources spaced 9.6nm. The convolution was subsequently normalized to a maximum of 40000 counts and a background of 20000 counts was added. The noise was finally added to this total image (source plus background). The synthesized image and the original sources are plotted in Figure \ref{fig:lineasoriginal} showing that the microscope would not resolve the two lines. 

\begin{figure}
\centering
 \includegraphics[width=0.8\linewidth]{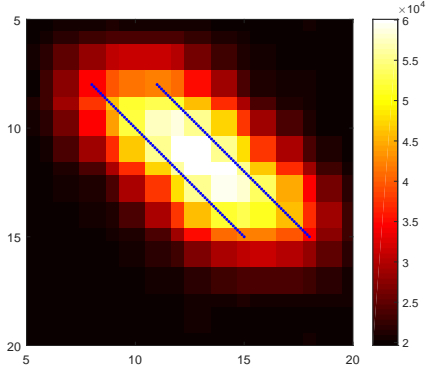}
\caption{Graphic of the source $R$ and the synthesized image $S$ obtained after  convolving $R$ with the IRF, adding the noise and the background. }\label{fig:lineasoriginal}
\end{figure}

Once the image was synthesized the algorithm was executed to obtain a preliminary result with an arbitrary number of virtual sources, in this case we use $N=600$.

 Observe that here we have that  $\tilde{I}=I$ so $G=0$ and we have almost all the terms needed to calculate $N_{op}$. Only the term $L$ remains (where we need explicitly $m$ and $R_p$). Here is where we use the first steps of the algorithm described in Section \ref{method}: we start with an arbitrary $N$ (in this case we use $N=600$), we find the solution for that $N$, then we make an histogram of the solution vector  for different bins  and define $m_{bin}$ (the number of nonzero bins).  Now we have an approximation of $y_p$, $R_p$ and $m$ so we can compute all the terms involved in $N_{op}$ (in this case we use $N=461$).

 Notice that for our reconstruction the number of virtual sources used does not need to match the actual number of sources used for the creation of the image. The technique intends to recover an approximate density, not to reproduce the exact solution.

\begin{figure}
\centering
\subfloat{%
       \includegraphics[width=0.5\linewidth]{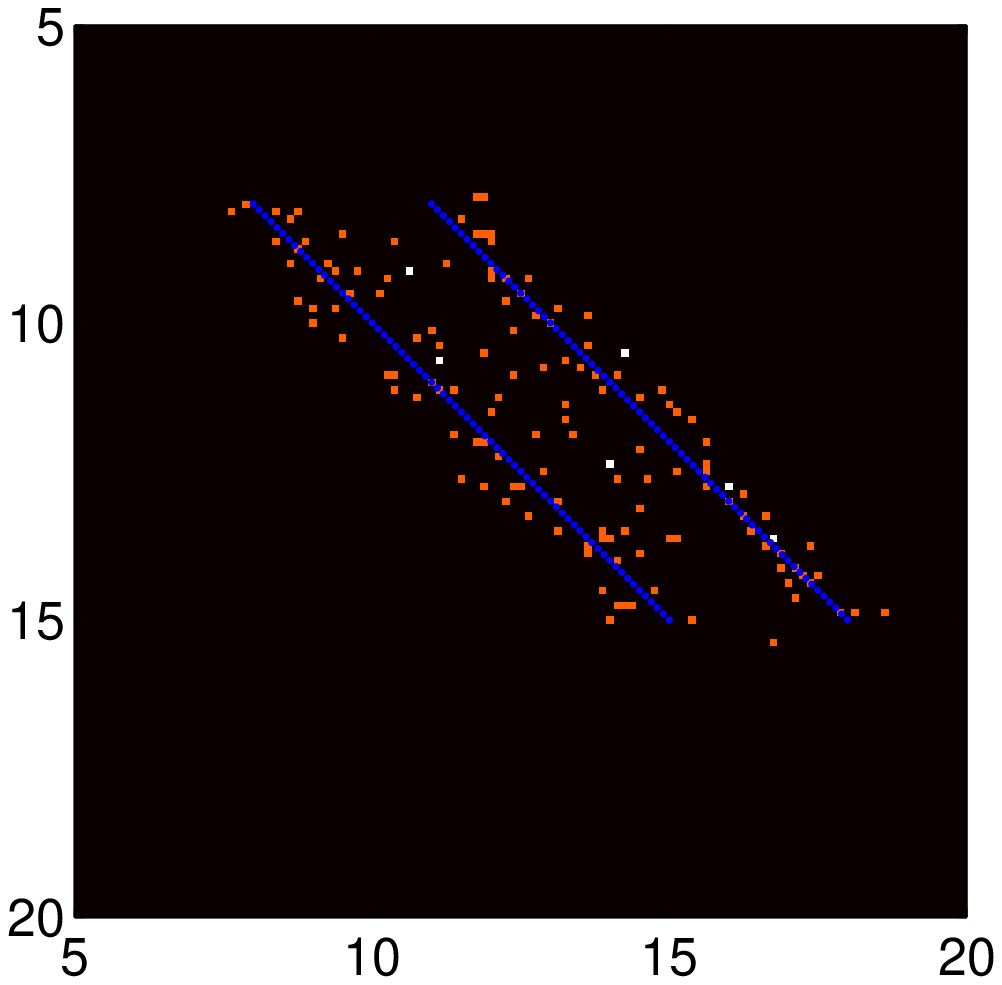}}
       \subfloat{%
        \includegraphics[width=0.37\linewidth]{{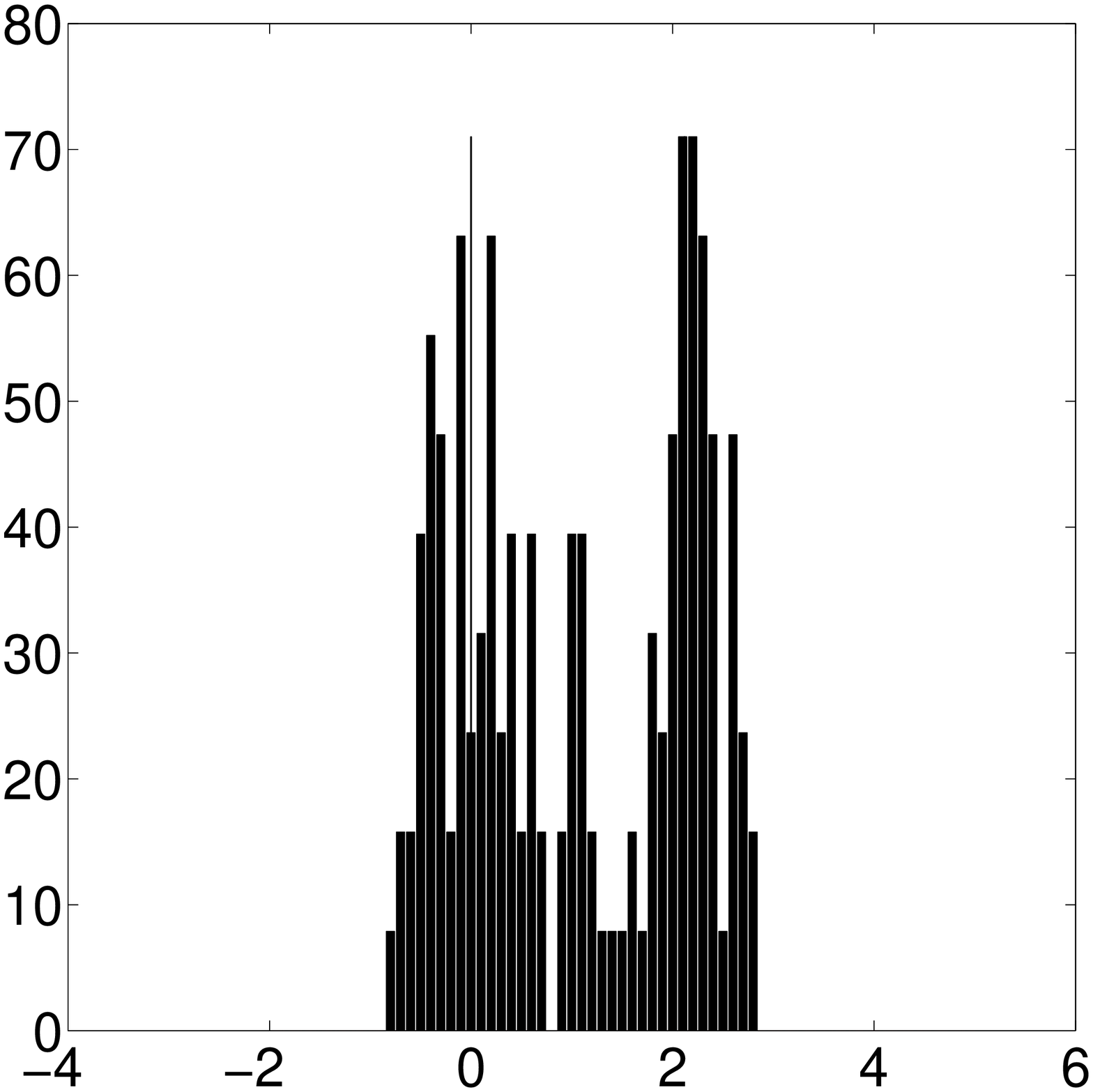}}}\\
        \subfloat{%
        \includegraphics[width=0.5\linewidth]{{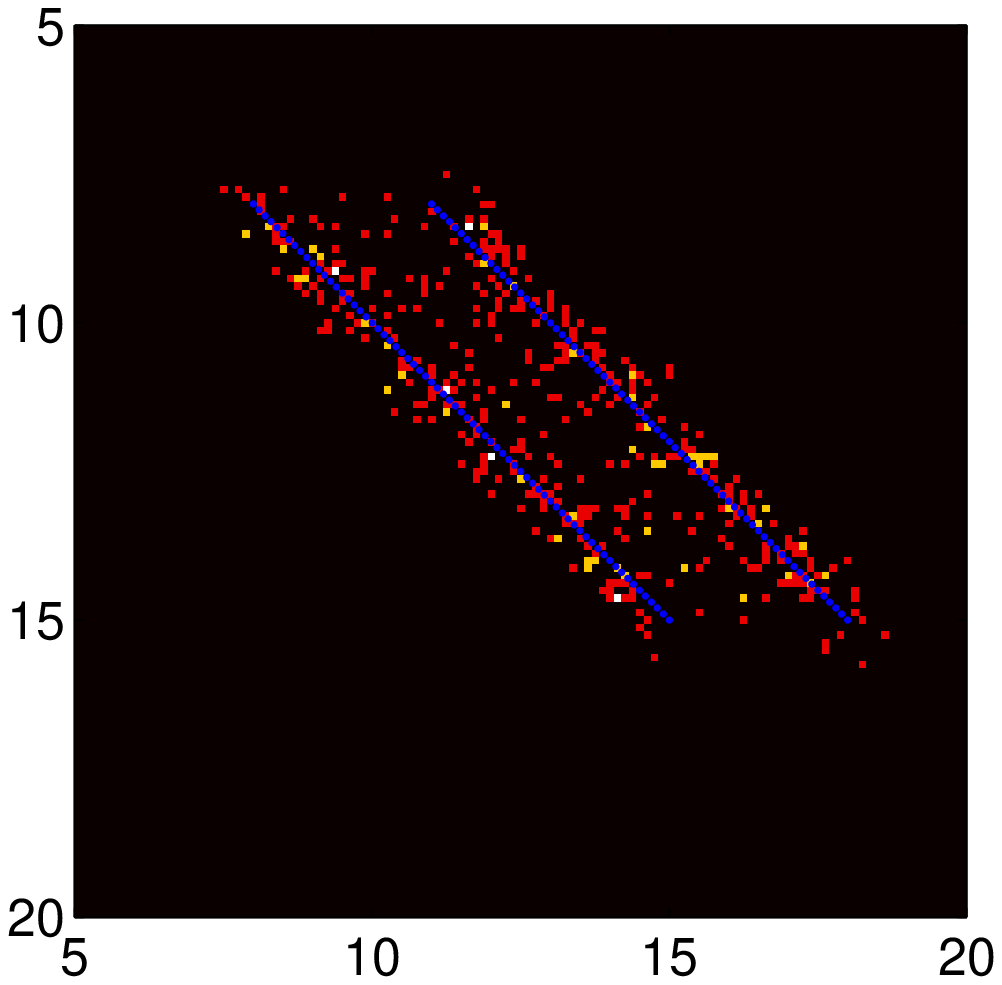}}}
        \subfloat{%
       \includegraphics[width=0.37\linewidth]{{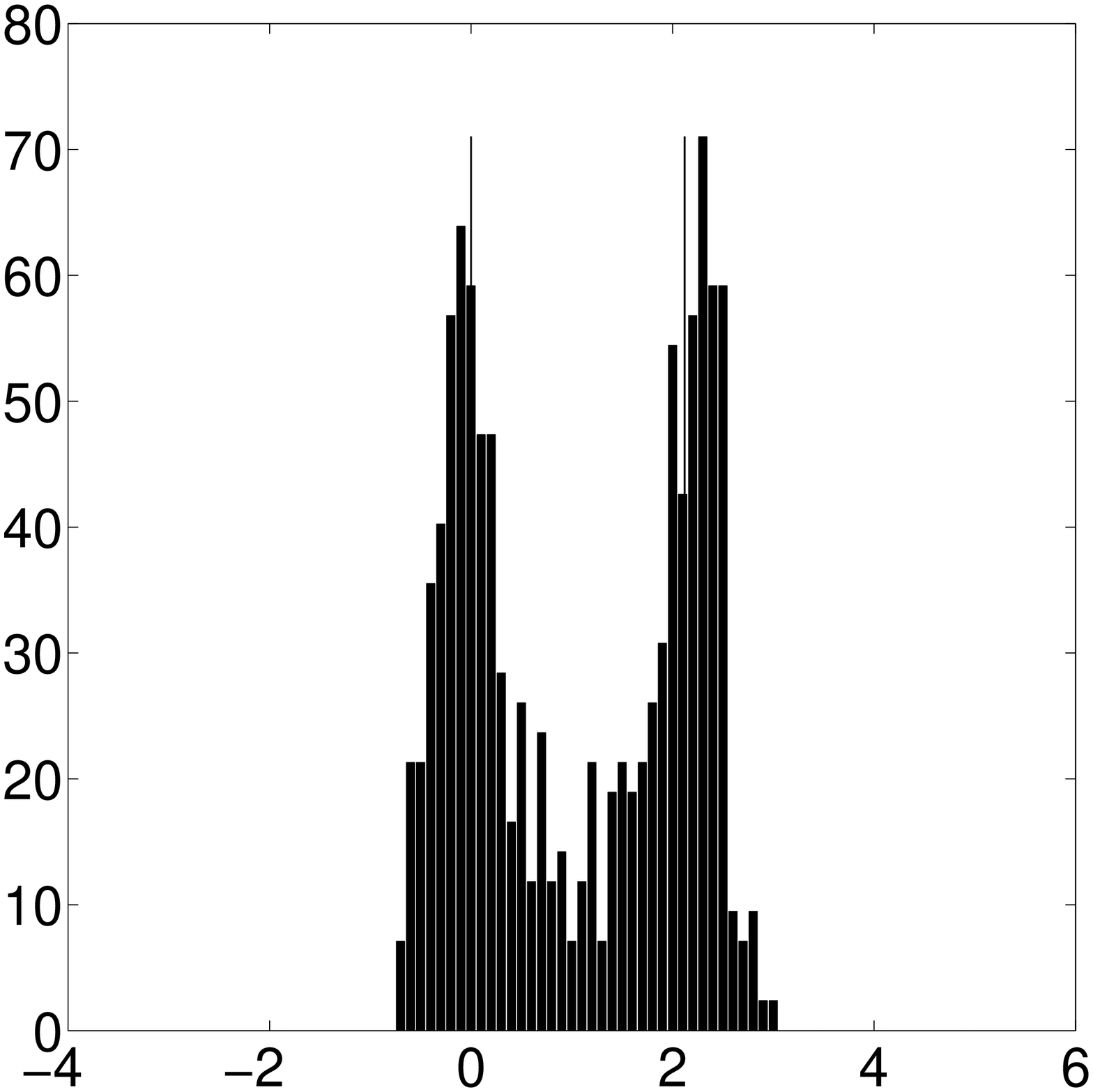}}}\\
       \subfloat{%
        \includegraphics[width=0.5\linewidth]{{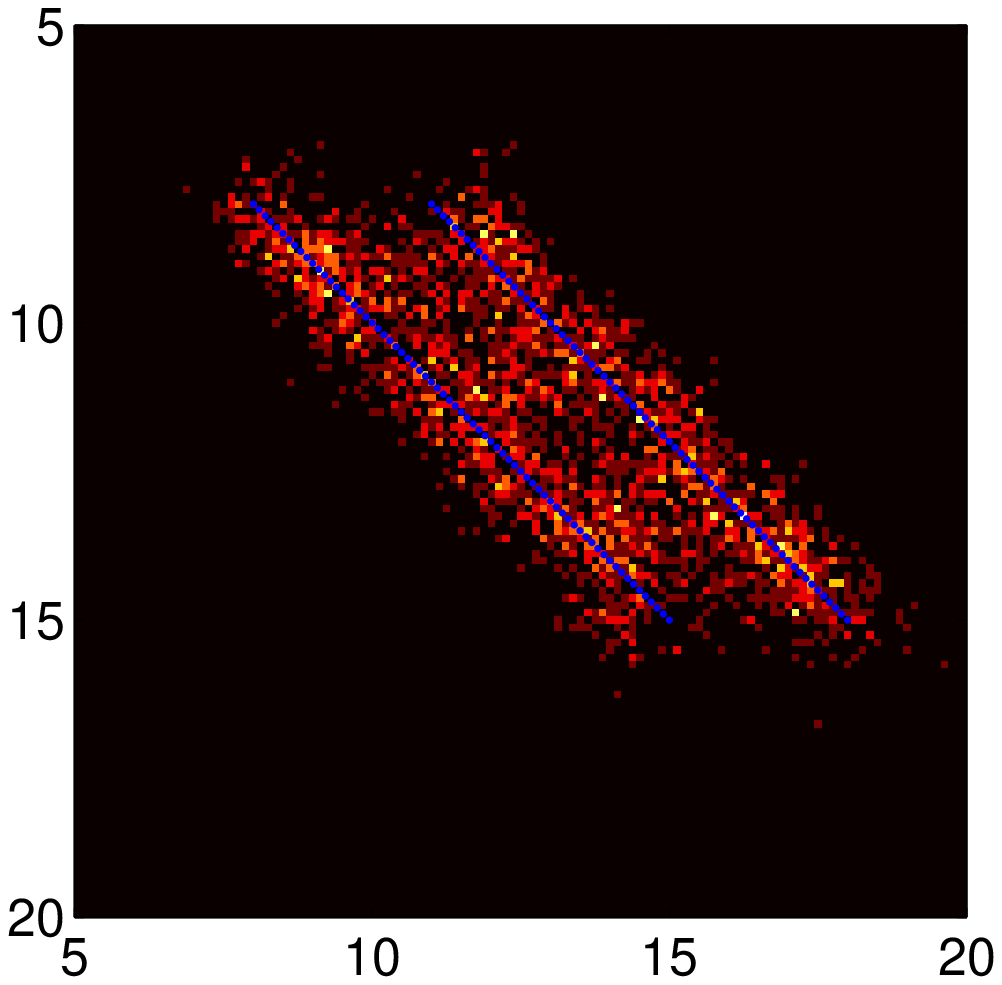}}}
        \subfloat{%
        \includegraphics[width=0.37\linewidth]{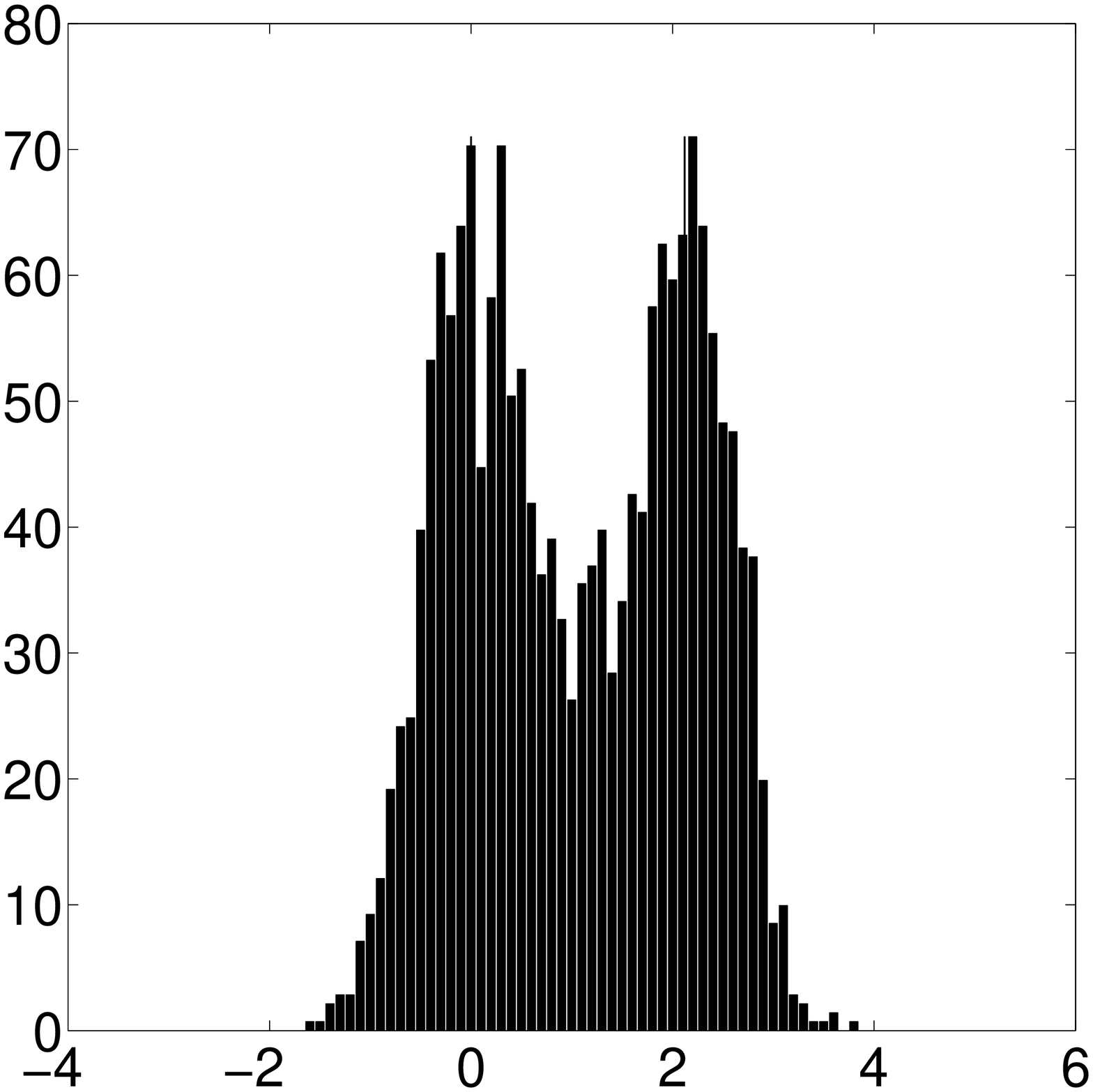}}            
        \caption{ First column: results using different number of virtual sources. In the first case $N=142$ which is equal to the actual number of sources, the second near the optimal $N=461$ and the last $N=2555$.   All the images are represented using  an histogram with a binning of $da=0.125$ pixel. Second column: histogram of the projection of the sources in the axis perpendicular to the lines of actual sources indicating the distribution of the solution around the gropund truth.}\label{fig:dosjemploslinea}
\end{figure}

%\begin{figure}\label{dosjemploslinea}
%\centering
%\subfloat{%
%       \includegraphics[width=0.5\linewidth]{recta142.eps}}
%       \subfloat{%
%        \includegraphics[width=0.5\linewidth]{{rectaN461.eps}}}
%        \subfloat{%
%        \includegraphics[width=0.5\linewidth]{{rectaN2555.eps}}}\\
%        \subfloat{%
%       \includegraphics[width=0.5\linewidth]{{hist142.eps}}}
%       \subfloat{%
%        \includegraphics[width=0.5\linewidth]{{hist461.eps}}}
%        \subfloat{%
%        \includegraphics[width=0.5\linewidth]{hist2555.eps}}            
%        \caption{ First line: results using different number of virtual sources. In the first case $N=142=$ which is equal to the number of original actual number, the second near the optimal $N=461$ and the last $N=2555$.   All the images are represented using  an histogram with a binning of $da=0.125$ pixel. Second line: The same as in the first line, here represented as an histogram of the projection in x after making a rotation of the axes.}
%\end{figure}

 In Figure \ref{fig:dosjemploslinea} the reconstruction for $N=2555$ and for the predicted optimum $N=461$ are presented. We also tested the case the number of virtual sources equals the number of actual sources, $N=142$. The predicted optimal $N$ resulted a better deconvolution as expected from the theoretical analysis. 
 To evaluate the precision in the fit the following strategy was used: project the positions in a new axis set such that the y axis is parallel to the lines and the x axis is perpendicular. Then run a histogram of the x projection and see how it groups around the position of the lines, i.e. $x=0$ for the first one and $x=  2.1214$ pixels for the second one. The histogram for the optimum $N=461$ shows a two lobe distribution with averages for each lobe departing less than $0.04$ pixels from the ground truth, and standard deviations of $0.4$ pixels $=27$ nm. This corresponds to a super-resolution factor $M_s=3.6$. As the number of sources is increased the lobes broaden slowly with $N$ as predicted. It can be seen from Figure  \ref{fig:dosjemploslinea} that for smaller or greater number of virtual sources, approaching the actual number, the solution gets worse. Hence the method is providing a good estimate of the optimal number of sources to be used to obtain the best spatial resolution.

The total number of sources that generated the image were 142 and the number of point sources within the IRF of the instrument are of the order of 40 (number of sources in a circle of diameter 2sigma). This indicates that a localization technique such a STORM or PALM would require more than a hundred frames to reconstruct the image and if a compressed sensing algorithm is used, as discussed in \cite{Candes2} and \cite{Min} for typical noise figures up to about 4 sources within the IRF can be recovered.

\end{exmp} 
 
 \medskip
\begin{exmp}
Real microscope image with unknown background.

\medskip
This example corresponds to the measurement of green fluorescent beads (520nm peak emission) under a microscope, capturing the signal with a CMOS camera. Here we have background fluorescence from the surrounding media that cannot be subtracted. In this example we consider $D=2$ and $B>0$. Each bead has a load of $10^4$ to $10^5$ fluorescent molecules (see \cite{Thermofisher}) and hence can be assumed a continuous distribution of sources. The relevance of this example is that here the IRF is not known and the error given by $g(x)$ must be estimated and that a reconstruction is made for a system with around $10^5$ sources within the IRF.

 The determination of a fit function for the IRF and the function $G$ is described in Apendix \ref{Fluorecencia}.    
In Figure \ref{tresejemplos} three different measurements and fits are shown. For the image acquisition an oil immersion objective was used with a total magnification to the camera of 96x. This yielded a pixel size of 67.7nm. The reconstructions of an 100nm isolated bead, a pair of 200nm beads not distinguishable in the original image and a cluster of 100nm beads are presented. In the original images the blur from the IRF is evident. The reconstructions show that for a single 100nm bead a region of scattered  sources with a radial standard deviation of the position of the sources of $\sigma=0.75$ pixels $=51$ nm  and corresponds to an improvement of a factor of more than 5 in the resolution of the instrument. In the absence of noise and $g=0$, as the bead is one used for the determination of the PSF,  the reconstruction should have yielded a single superpixel, and the scattering is an indication of the uncertainty of the reconstruction.  

To estimate $\sigma_{op}$, in this case we start with $N=160$ we make the histogram for different values of $da$ and we finally arrive to $\sigma_{op}= 1.2$ pixels  (81nm) and with $N_{op}=45$. This is in fact a very good estimate if we compare it with the radial standard deviation of the particles mentioned before. 

The blurred image of the 200nm beads reconstructs to two beads 200nm apart, consistent with two contacting beads. In this case we obtain a $\sigma_{op}=1.5$ pixels$=100$nm which again is an overestimation of a factor around 2 when compared with the actual data, indicating that for dense structures a better reconstruction than predicted can be expected.

The third case is a cluster of 100nm beads confirming the high spatial resolution obtained.

% \begin{figure}
%    \centering
%    \includegraphics[width=3.5in]{solutions}
%      \caption{Results for a source of 100 nm. We represent the original image  the solution for an histogram with $da=0.5$ and then we represent the solution convolved with an shpere of 100nm.}\label{fig:Sol}
%\end{figure} 

\begin{figure}\label{tresejemplos}
\centering
\subfloat{%
       \includegraphics[width=0.35\linewidth]{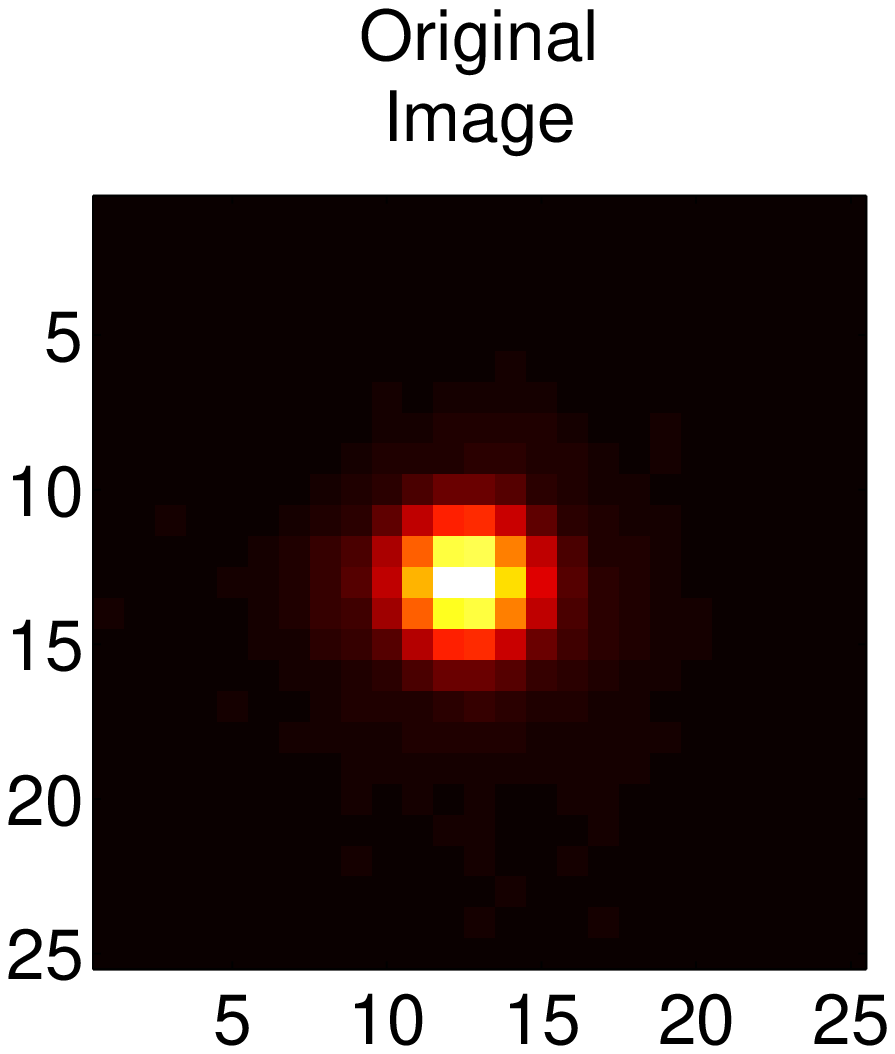}}
       \subfloat{%
        \includegraphics[width=0.35\linewidth]{{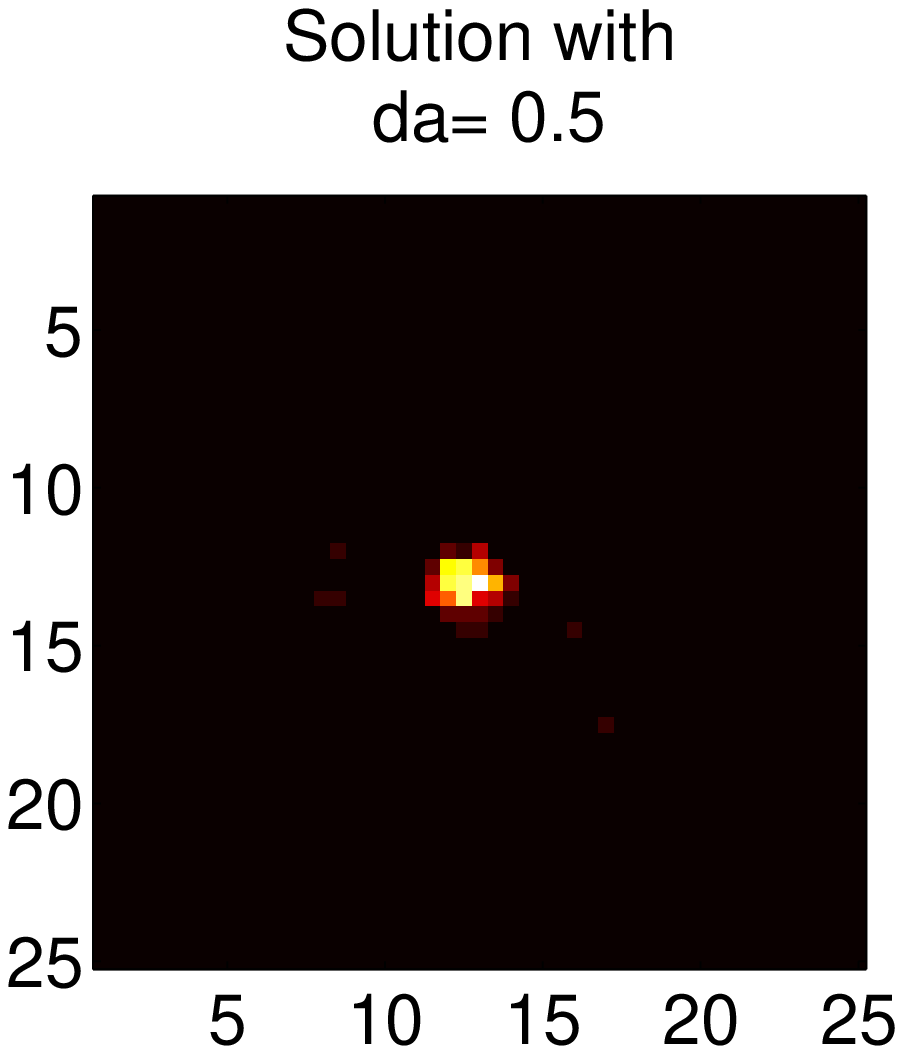}}}
        \subfloat{%
        \includegraphics[width=0.35\linewidth]{{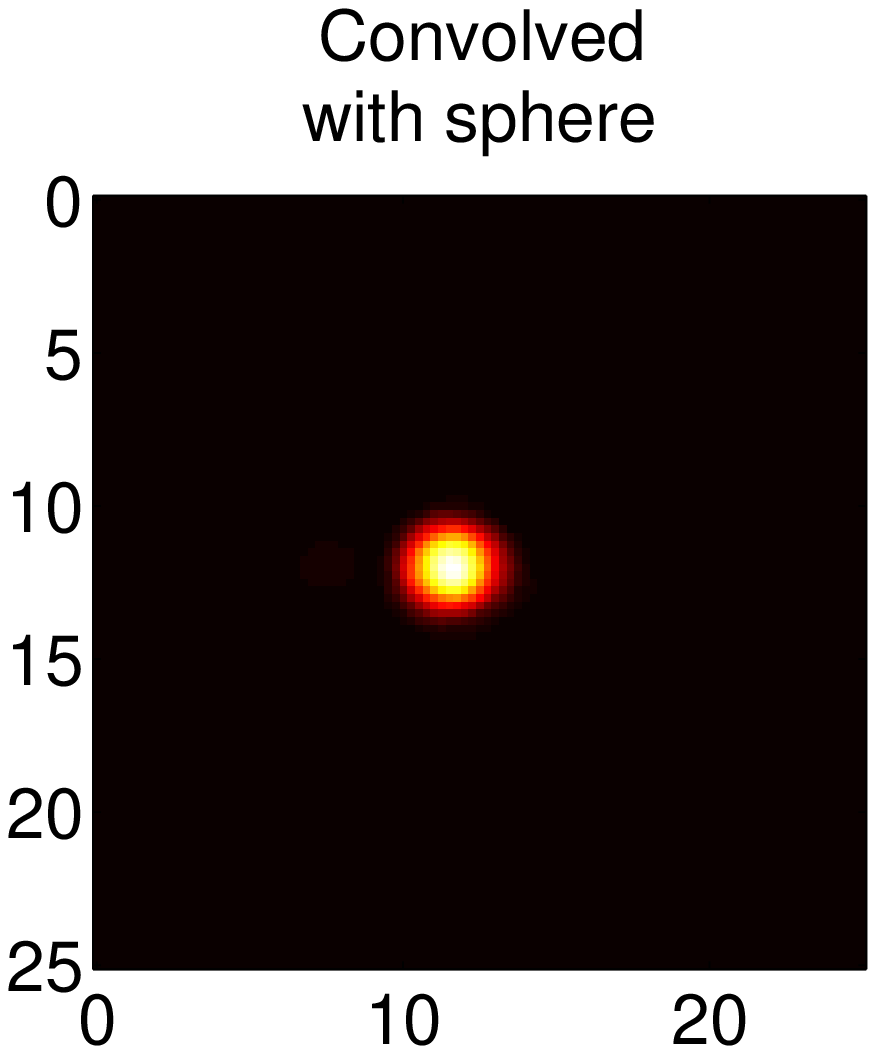}}}\\
        \subfloat{%
       \includegraphics[width=0.35\linewidth]{{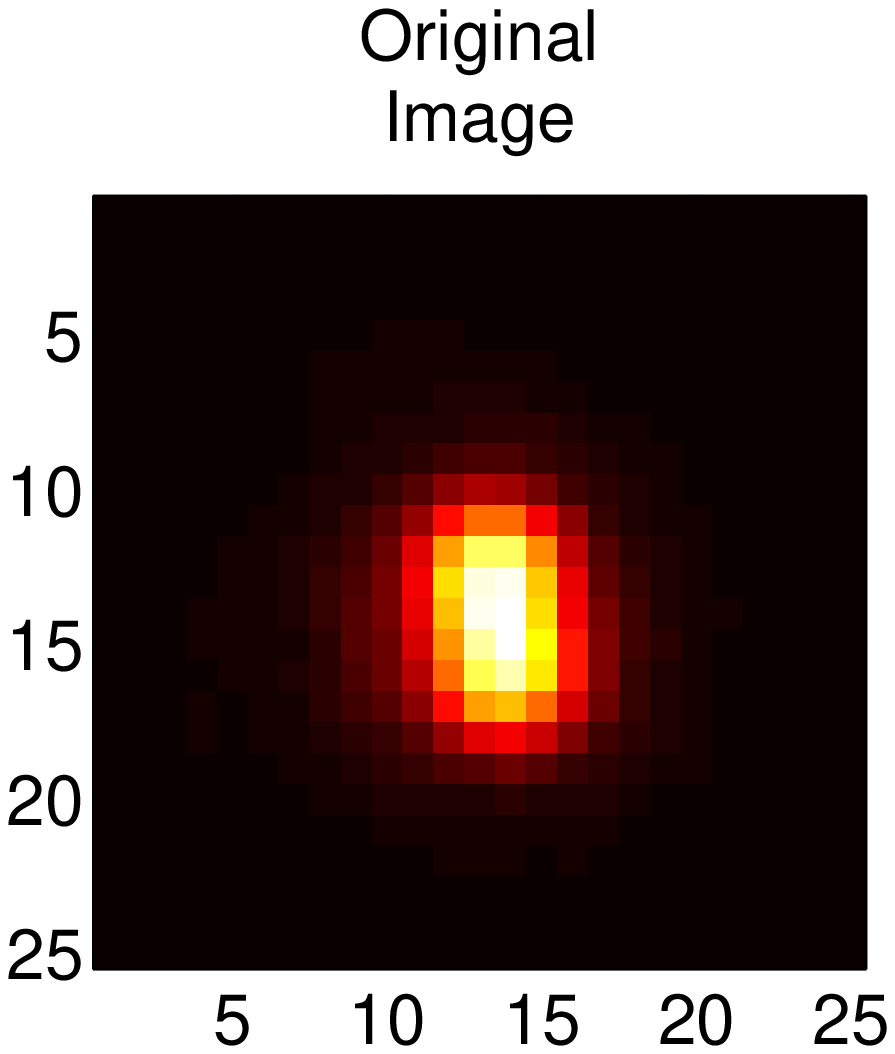}}}
       \subfloat{%
        \includegraphics[width=0.35\linewidth]{{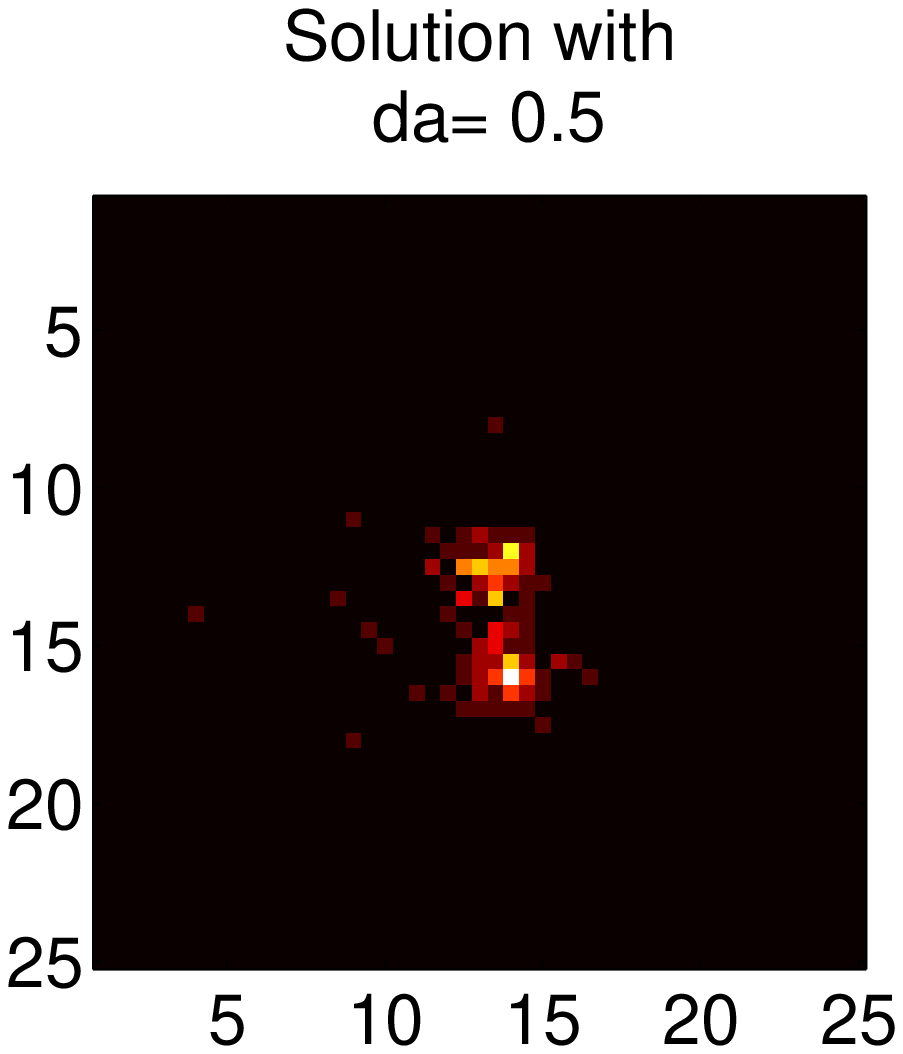}}}
        \subfloat{%
        \includegraphics[width=0.35\linewidth]{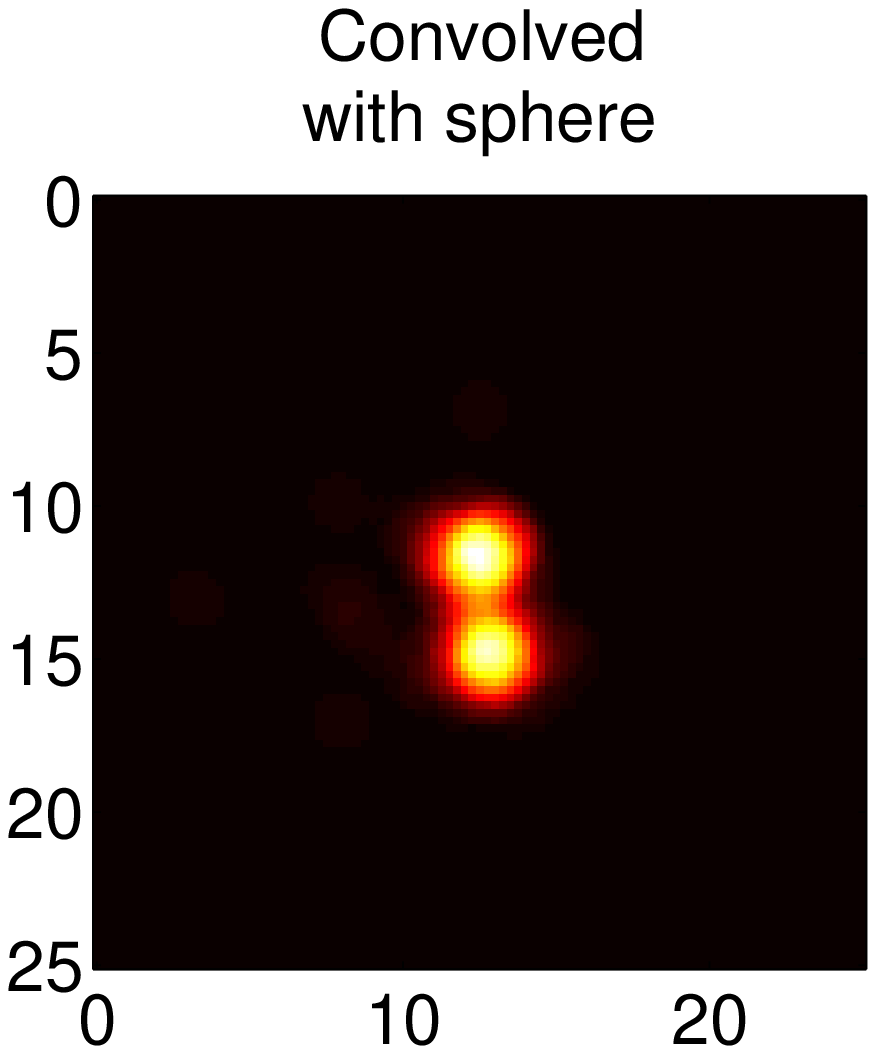}}\\
        \subfloat{%
       \includegraphics[width=0.35\linewidth]{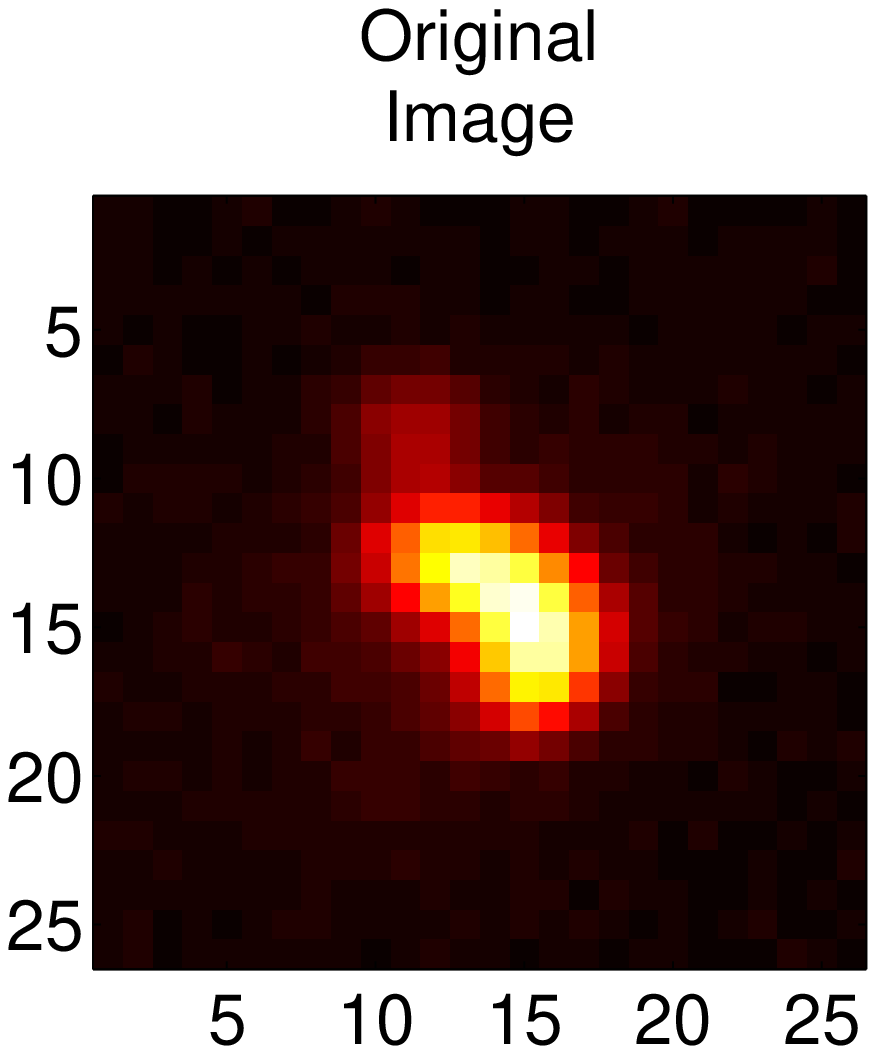}}
       \subfloat{%
        \includegraphics[width=0.35\linewidth]{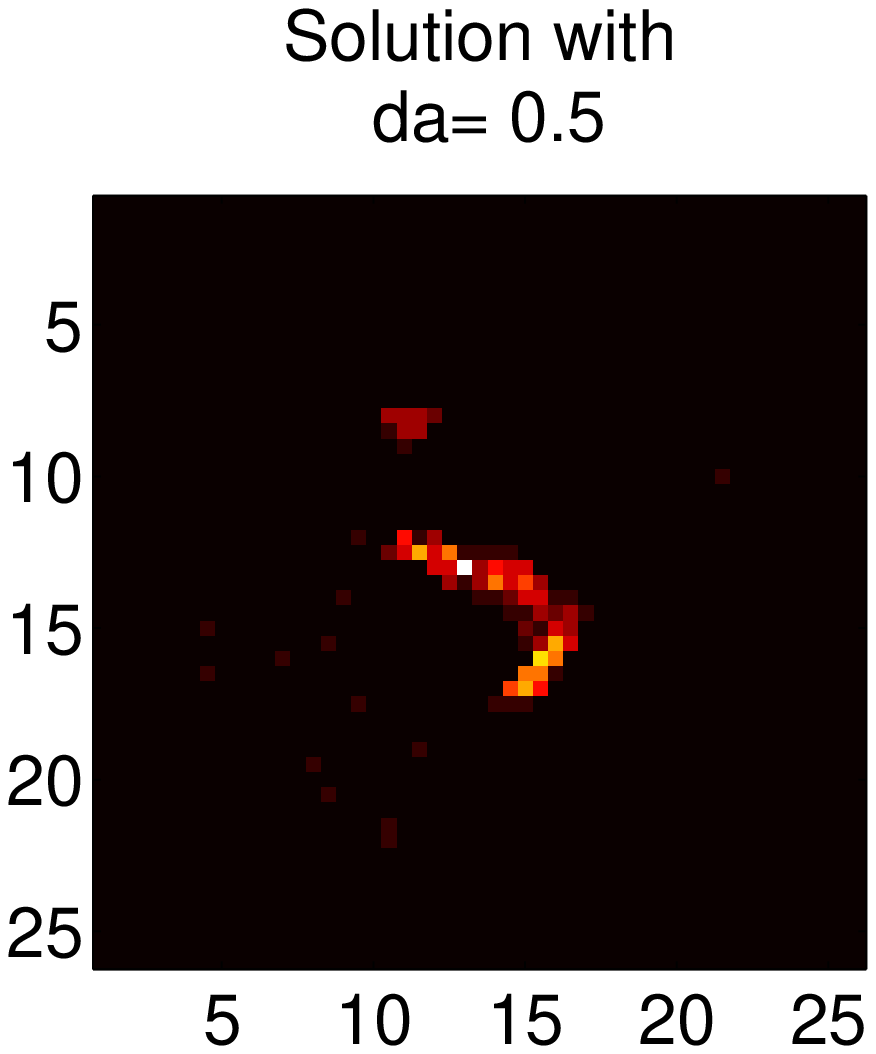}}
        \subfloat{%
        \includegraphics[width=0.35\linewidth]{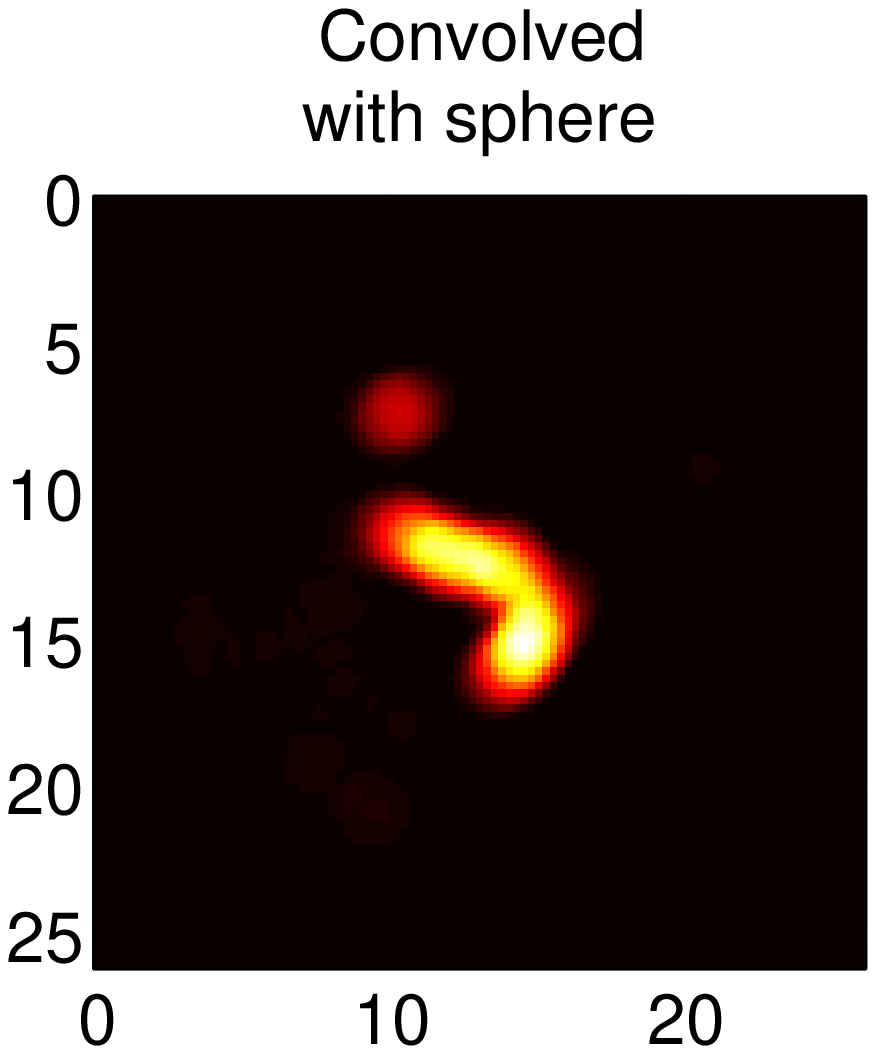}}
         \caption{Top to bottom: results for a source of 100 nm, cluster of 200nm and a cluster of 100nm. Left to right: We represent the original image  the solution for an histogram with $da=0.5$ and the solution convolved with a 100nm sphere.}
\end{figure}

   \end{exmp}
\section{Notation}

\label{sec.tablas} 
 
 \begin{table}[H]	
\begin{tabular}{|c|l|} \hline  
$D$ & Dimension of the space.	 				\\
		$x_i$ &  Pixel $i=1\dots n$.	 				\\
		$S(x_i)$    &  Measurement.	\\  
		$R(x)$  & Density of the source (unknown).
			\\
			$R_p$  &  Intensity of the point sources (unknown).
			\\
		$y_p$   & 	Position of the  $m$  point sources (unknown). 					\\
		$\eta(x_i)$   &  Noise.			\\
		$d_0$ & 2 \mbox{Standard deviation of } $I(x)$. \\ 
		$d_p$ & {size of the pixel}.\\ 
		$d_s$ & {size of the superpixel}.\\
		$\sigma$ & Uncertainty in the positions.\\ 
		%$M_p$& $\frac{d_0}{d_p}$ \\
		$M_s$ & Superesolution factor: $\frac{d_0}{ 2 \sigma }$.\\
		 						$I(x)$  &  IRF  (convolved with the pixel). 				\\
		$\tilde{I}(x_i)$  &  Approximation of $I$. \\
		$N$  & 	 Number of sources 	used for the fit.			\\
		$\alpha$  &  Intensity of the each source.\\ 
		$a_k$   & Position of the $N$ point sources (with repetitions).			\\
		$\tilde{a_k}$  &  Approximation of $a_k$. Minimizer of the problem.	 				\\
		$N_p$ & Rounding to the nearest integer of $\frac{R_p}{\alpha}$. \\
		$\bar{R}(x)$ & Truncation of $R(x)$: $R_p=N_p \alpha$.\\
		$\tilde{R}(x)$ & Minimizer of the problem.\\
		\hline	 
		\end{tabular}
\caption{Table of definitions} \label{tab:def}
\end{table}

\section{Conclusion}
A new method for super-resolution method for data deconvolution  from a single acquisition has been presented. The method relies in approximating the solution by a superposition of point sources of fixed amplitude (SUPPOSe). In this manner the problem of finding a positive value for the intensity at each pixel is converted to finding the position of the sources, which is an unconstrained problem. A minimization problem results that due to the large dimension of the space (coordinates of the sources) must be solved in a global manner. A genetic algorithm was chosen for this purpose. By construction the solution is  positive, and the method avoids the amplification of noise that appear in Fourier based techniques, which would result in limitations in the spatial resolution. An upper bound for the uncertainty in the position of the sources was derived and two very different experimental situations were used for the test as well as an artificially generated image  showing excellent reconstructions and  that  the method always performed a factor of 2 to 3 better than the predicted bound. The method also provides a way to determine the optimum number of sources to be used for the fit. The method requires a certain degree of sparcity, working better for sparcer sources, but the examples presented where order of magnitude denser than those that can be resolved with prior super-resolution techniques in a single acquisition. Examples with more than $10^4$ sources within the IRF were shown. For the reconstruction of fluorescent microscope images a resolution of $\lambda/10$ was demonstrated.

As a final remark it is worthwhile noticing that as formulated the algorithm can also be applied for cases where the IRF is not translational invariant, situation in which Fourier based methods are not applicable. Another extension of the method would be to nonlinear expressions linking the point source emission to the detected signal.

\appendix%[Supplementary Material ]

\subsection{Fitting of the Instrument response function}\label{fitting}

The determination of the instrument response function was made  with the following procedure:
\begin{itemize}
	\item
Acquire the data for known point like sources.
\item Fit each source with a tailored function that resembles the shape of the data obtained.
\item Shift all the point like sources to a common origin using the fit obtained individually. And normalize them to the same area.
\item Fit the complete set of co-centered point sources.
\item Determine the residue as an estimate of the error $g(x)$
\item Determine the autocorrelation of the residue $G(x).$
\end{itemize}

\subsubsection{Spectral lines}\label{espectro}

For the calibration we chose three isolated peaks that from the NIST table (see \cite{NIST})  corresponding to  Hg $546.07350$  nm, Kr $587.09$nm  and He $587.56$nm. We use these three peaks $S_r$ $r=1,...,3$ to fit the function $I$.  We also use these peaks to estimate the function $g$ and  the calibration  wavelength vs pixel.

We call $x_r$ each vector (zone) where we take each $S_r$ (may not have the same size).
We normalize and co-center each peak by: 
$\bar{S}_r=\frac{S_r}{\sum S_r}$, 
$\bar{x}_r=x_r-\bar{S}_r\sum x_r$
so all the peaks are centered around zero and with sum equal one. Here the sum is taken over all the  pixels of $S_r$.

We finally fit the points $[\bar{x}_1\ \bar{x}_2\ \bar{x}_3]$ and $[\bar{S}_1\ \bar{S}_2\ \bar{S}_3]$ to fit the function $\tilde{I}$ by an asymmetric function of the form $$\tilde{I}(x)= \frac{a_1}{e^{b_1x}+e^{-{b_2x}}}.$$
where a value of $d_0= 5.6$ pixels $=1.23$ nm was obtained.

\begin{figure}
\includegraphics[width=1\linewidth]{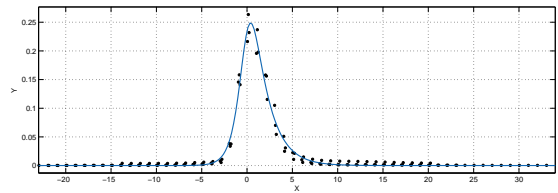}
\caption{Shifted and normalized data for the 3 data and the function $I$.}\label{fig:I03datos}
\end{figure}

Then we define,
$$p(j) =\mbox{ 
closest pixel to the center of each } \bar{S}_r $$
 and the function $g$ can be approximated by 
$$    g(p)=\frac{1}{3}\sum_{r=1}^3(\bar{S}_r(p(r))-\tilde{I}(\bar{x}_r(p(r)))).
$$

The function selected for the fit was a result of several trials with asymmetric functions until an adequate bound for g was achieved. In this manner g resulted small enough to make its contribution to the uncertainties negligible when compared to those arising from the noise of the measurement in (21).

Finally to compare the results we make an histogram with a binning of $d_a=0.5$ pixel $=0.11 \mbox{nm}$ and  we use the calibration to convert the results from pixels to wavelength. 
\subsubsection{Point spread function of the microscope}\label{Fluorecencia}

As point surces fluorescent beads 50nm nominal radius were used. They were mounted on a microscope slide and embedded in an antifading agent. Images were obtained that contained many beads. A program automatically selected the bright spots and selected a small region around the identified maximum. 

After fitting all the available beads with adequate brightness, the range of acceptable widths was determined such that cases with multiple beads within the fitting area were discarded.

To approximate the function $\tilde{I}$ we used $42$ of such single bead images ($S_r$). Each $S_r$ can be expressed as \eqref{partida} by
$$S_r(x)=I(x)+\eta_r(x)+B_r(x),$$
 for some background $B_r$ and $\eta_r$ the noise. We fit this equation using a adequate fitting function plus a constant (the background). We note $J_r$ to each fitting function where 
   $J_r$ has the form $J_r(x)=I_r(x)+a_r$. 
   We use these individual fits to normalize and co-center all the measurements of individual beads, we call these $S'_r$. Finally we  make a single fit with all the data, generating in this manner the function $\tilde{I}(x)$.

If we assume that,
 $\langle B_r \rangle =\langle a_r \rangle$
 and make the following approximation of the function $g(x)$, 
  \begin{equation}g(x)=\sum_{r=1}^s  \frac{(S'_r-\tilde{I})(x)}{s}.\label{residuo}\end{equation}
  So in this case we have an explicit formulation of the function $g$, for each pixel.
  The first trials with Gaussian functions and with theoretical predictions for the PSF did not yield a satisfactory value for $g$ and the final choice was the following function
  $$\tilde{I}(x)=b_1e^{-\rho^2d_1}+b_2\rho^2e^{-(\rho-\rho_0)^2d_2}.$$
     where $\rho=\|x\|.$ This function is dominated by a centered Gaussian plus a halo centered at $\rho_0=3.9$  pixels$=264$nm. The resolution of this PSF is characterized by $d_0=3.94$ pixels$=265$nm.

\begin{figure}
\includegraphics[width=1\linewidth]{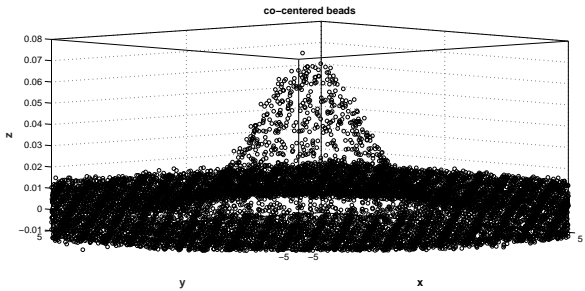}
\caption{Shifted and normalized data for the $42$ beads used.}\label{fig:cocenteredbeads}
\end{figure}

\begin{figure}
\centering
\subfloat{%
        \includegraphics[width=0.5\linewidth]{{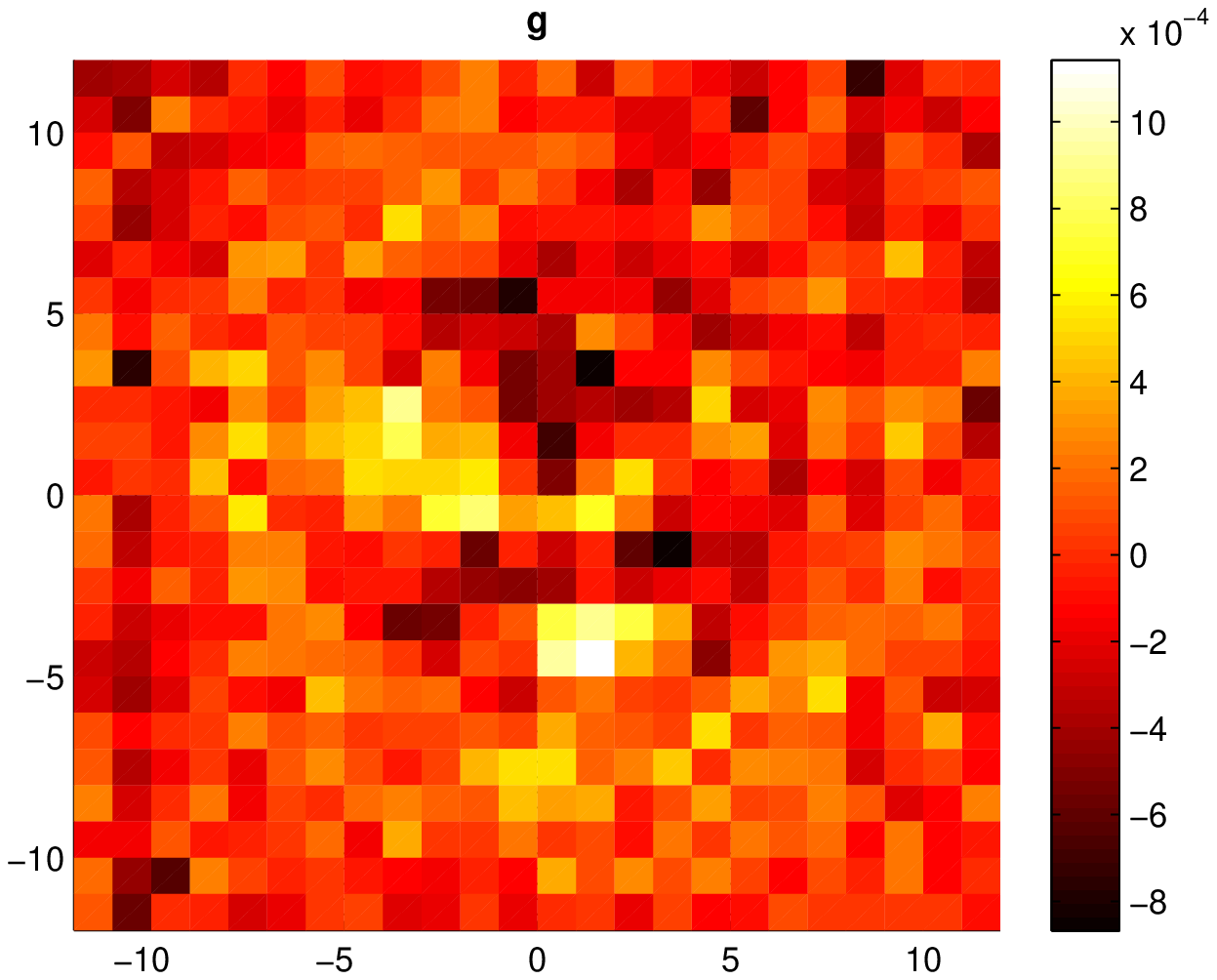}}}
        \subfloat{%
        \includegraphics[width=0.5\linewidth]{{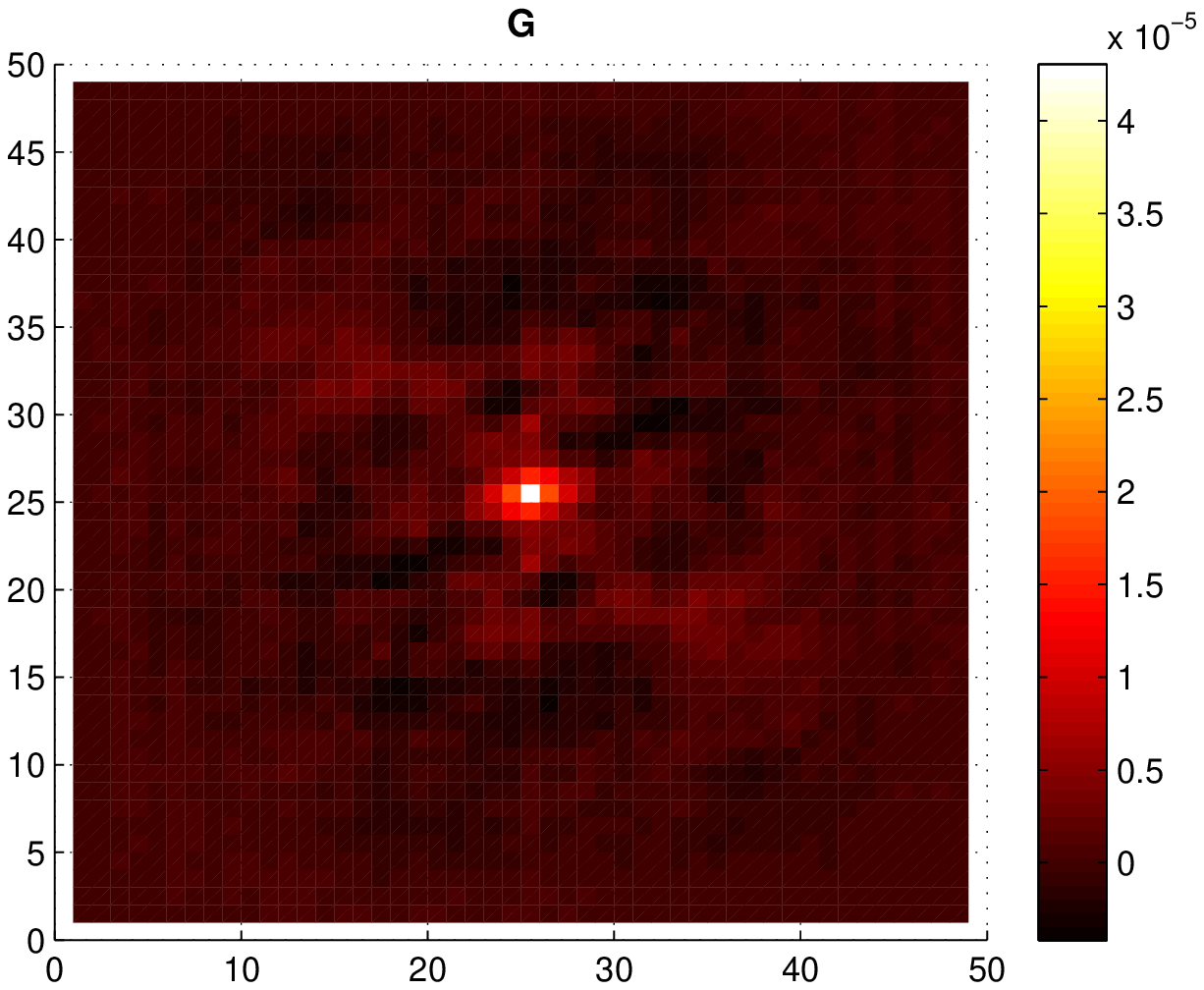}}}
                         \caption{Function $g$ with a maximum of order $10^{-4}$ and $G$  with a maximum of order $10^{-5}$.}\label{fig:funciongG}
\end{figure}

 Figure \ref{fig:cocenteredbeads} shows the shifted and normalized data for the $42$ beads used. Figure \ref{fig:funciongG} shows the residue $g(x)$ as obtained from \eqref{residuo} and its autocorrelation $G(x)$.

\subsection{The genetic algorithm}

Each generation consisted of $M$ individuals, being an individual a vector with the coordinates of the $N$ sources. As fitness function to be maximized the inverse of the $\chi^2$ with an offset to increase the difference between individuals was used. 

\subsubsection{The Initial Family}

Mimicking steps (1), (2) and (3) of the algorithm to find $N$ of Subsection \ref{background} we can obtain a initial family $a_{kl}\in\R^D$ where $l=1,..,M$ and $k=1,...N$ (that is we have $D$ matrices in $\R^{M\times N}$). The only difference here is that we have to change step (2) making a random perturbation of the maximum to generate the $M$ different individuals  of the family, that is now step (2) becomes:
\begin{align}\tag{2'}\begin{split}
a^{i}_{kl}&=(b^{i}_k+r^i)\quad \mbox{for all } i=1,\cdots, D
\\ T(x)&=T(x)-\alpha_0 \sum_{k=1}^{i} \tilde{I}_{dev}(x-(b_k+r))\end{split}\end{align}
where $r\in \R^D$ is a random vector with normal distribution with standard deviation proportional to de IRF width (this $r$  changes as we go through all the values $k=1,..,N$ and $l=1,..,M$).

\subsubsection{One iteration of the Algorithm}

In each generation a fraction of the best individuals was kept unmodified and then the full family was modified with the following sequence:

\begin{enumerate}
\item The best fitted $ne$ individuals (Elite) are saved. That is, the individuals with smaller $\chi^2$.
\item
The best individuals were duplicated as many times as the integer part of the fitness function.
\item
The new family was completed up to $M$ with the following best fit individuals.
\item An amount $p$ of the individuals selected randomly was crossed by exchanging between pairs with probability $1/2$ the coordinates of the sources.
\item A  number $n$ of the individuals selected randomly suffered mutations.
Also only a  fraction of the coordinates sources call $pormut$ selected randomly suffered mutations. These mutations consist  in shifting the position of the sources by a random fraction of $\rho_0=\sigma parmut $ where $\sigma$ is the IRF width and $parmut$ is a parameter of the algorithm.
\item In the case there is background once we have found $\{\tilde{a_k}\}_{k=1}^N$ we use a linear Least Squares fit to find a corrected value for $\alpha$. 
\end{enumerate}

With this new generation the procedure was repeated until the $\chi^2$ reached the theoretical minimum due to noise or does not improve any more or after a predefined number of generations.

\begin{figure}
\centering
       \includegraphics[width=0.8\linewidth]{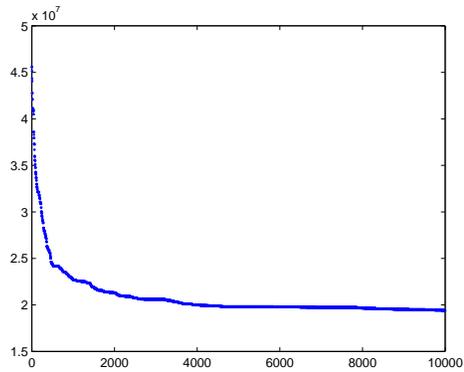}
\caption{Evolution of $\chi^2$. In this case the synthetic Noise satisfy $\|\eta\|^2=1.91 \times 10^7$ and the function $\chi^2$ after $10^4$ iterations is equal to $1.93\times 10^7$.}
\label{fig:graficochi}       
       \end{figure}

In Figure \ref{fig:graficochi} we can see how the $\chi^2$ evolves after $10^4$ iterations of the algorithm. In this case the algorithm was applied to the synthetic image of Figure \ref{fig:lineasoriginal}.

%\bibliographystyle{IEEEtran}
%\bibliography{IEEEabrv,mybibfile}
%\bibliography{IEEEabrv,particulasbiblio}

% or
%\appendix  % for no appendix heading
% do not use \section anymore after \appendix, only \section*
% is possibly needed

% use appendices with more than one appendix
% then use \section to start each appendix
% you must declare a \section before using any
% \subsection or using \label (\appendices by itself
% starts a section numbered zero.)
%
%\cite{IEEEexample:incollectionmanyauthors}

%\cite{IEEEexample:articleetal}

%\cite{IEEEexample:article_typical}

% you can choose not to have a title for an appendix
% if you want by leaving the argument blank

% use section* for acknowledgment
%\section*{Acknowledgment}

%The authors would like to thank...

% Can use something like this to put references on a page
% by themselves when using endfloat and the captionsoff option.
\ifCLASSOPTIONcaptionsoff
  \newpage
\fi

\begin{IEEEbiographynophoto}{Sandra Mart\'inez}
Received the Ph.D. degree in the Department of Mathematics of the University of  Buenos Aires, Argentina in 2007. She is currently a Professor with the University of  Buenos Aires and researcher of the National Research Council (CONICET).  Her current research interests include super-resolution problems and numerical methods for partial differential equations.
\end{IEEEbiographynophoto}

%\vspace{-10cm}

\begin{IEEEbiographynophoto}{Oscar E. Mart\'inez}
Born in Buenos Aires on 1953. Ph.D. in Physics (UBA-1982). Bell Laboratories 1982-1984. MTS of CITEFA until 1986, CNEA until 1993 and full professor at UBA (Universidad de Buenos Aires). Co-founder of the startup company Tolket SRL. Former Fellow of the OSA. Member of the Staff of the CONICET (1985-) and associate member of the ICTP (UN)  2003-2007. Has published more than 100 papers with more than 2000 citations and filed 10 patents. Area of expertise: ultrafast lasers, near field optics, novel optical instrumentation. Areas of current research: photothermal phenomena for materials science, nano-optics and biophtonics.

\end{IEEEbiographynophoto}

% insert where needed to balance the two columns on the last page with
% biographies
%\newpage

%\begin{IEEEbiographynophoto}{Jane Doe}
%Biography text here.
%\end{IEEEbiographynophoto}

% You can push biographies down or up by placing
% a \vfill before or after them. The appropriate
% use of \vfill depends on what kind of text is
% on the last page and whether or not the columns
% are being equalized.

%\vfill

% Can be used to pull up biographies so that the bottom of the last one
% is flush with the other column.
%\enlargethispage{-5in}

% that's all folks
\end{document}